\documentclass{article}

\usepackage[utf8]{inputenc} 
\usepackage[T1]{fontenc}    
\usepackage{hyperref}       
\usepackage{url}            
\usepackage{booktabs}       
\usepackage{amsfonts}       
\usepackage{nicefrac}       
\usepackage{microtype}      
\usepackage{xcolor}
\usepackage[table]{xcolor}
\usepackage{graphicx}
\usepackage{subcaption}
\usepackage{caption}
\usepackage{hyphenat}
\usepackage{setspace}
\usepackage{amsmath}
\usepackage{amssymb}
\usepackage{mathtools}
\usepackage{amsthm}
\usepackage{longtable}
\usepackage{multirow}
\usepackage{float}
\usepackage{tabularx}
\usepackage{enumitem}
\usepackage{threeparttable}
\usepackage{algorithm}
\usepackage{algorithmic}
\usepackage{titletoc}
\usepackage[capitalise]{cleveref}
\usepackage{wrapfig}
\usepackage{makecell}

\usepackage[preprint]{my_style}

\usepackage{tcolorbox}
\tcbuselibrary{skins, breakable, listings}

\definecolor{tabcolor}{rgb}{0.64, 0.87, 0.93}
\definecolor{capcolor}{rgb}{0.55, 0.80, 0.98}
\definecolor{brandblue}{RGB}{57,95,207}
\definecolor{linkblue}{HTML}{0064E0}
\definecolor{textgray}{HTML}{1C2B33}
\definecolor{boxbg}{HTML}{F1F4F7}

\hypersetup{
  colorlinks=true,
  linkcolor=brandblue,
  citecolor=brandblue,
  urlcolor=linkblue
}

\newcommand{\paperTitle}{Proxy OPD: On-Policy Distillation with Transferable Relative Proxy Update for Large Language Models}

\newcommand{\paperAuthors}{%
  {\sffamily\bfseries Daocheng Fu$^{*,1,2}$}, 
  {\sffamily\bfseries Rong Wu$^{*,1,3}$}, 
  {\sffamily\bfseries Yu Yang$^{*,1,3}$},
  {\sffamily\bfseries Jianbiao Mei$^{1}$},
  {\sffamily\bfseries Licheng Wen$^{4,5}$},\\
  {\sffamily\bfseries Xuemeng Yang$^{\dag,1}$},
  {\sffamily\bfseries Yong Liu$^{3}$},
  {\sffamily\bfseries Botian Shi$^{1}$},
  {\sffamily\bfseries Yu Qiao$^{1}$}
  {\sffamily\bfseries Pinlong Cai$^{\dag,1}$},
}

\newcommand{\paperAffiliations}{%
  {\normalsize $^1$ Shanghai AI Laboratory}, 
  {\normalsize $^2$ Fudan University},
  {\normalsize $^3$ Zhejiang University},\\
  {\normalsize $^4$ Shanghai Innovation Institute},
  {\normalsize $^5$ Shanghai Jiao Tong University}
}

\newcommand{\paperNotes}{%
  {\small $^*$ Equal Contribution}, {\small $^\dag$ Corresponding Author}%
}

\newcommand{\githubLink}{\url{https://github.com/KnowledgeXLab/P-OPD}}
\newcommand{\huggingfaceLink}{\url{https://huggingface.co/KnowledgeXLab/P-OPD_Experiments}}
\newcommand{\publishDate}{\today}

\newcommand{\renderFrontBox}{%
    \tcbset{
    enhanced, frame hidden,
    colback=boxbg,
    left=0.5cm, right=0.5cm, top=0.5cm, bottom=0.5cm,
    arc=16pt,
    before skip=0pt,
    grow to left by=1.5pt, grow to right by=1.5pt,
    overlay={
    \node[anchor=north east, at=(frame.north east), xshift=-2.3cm, yshift=-0.45cm] 
        {\includegraphics[height=1.00cm]{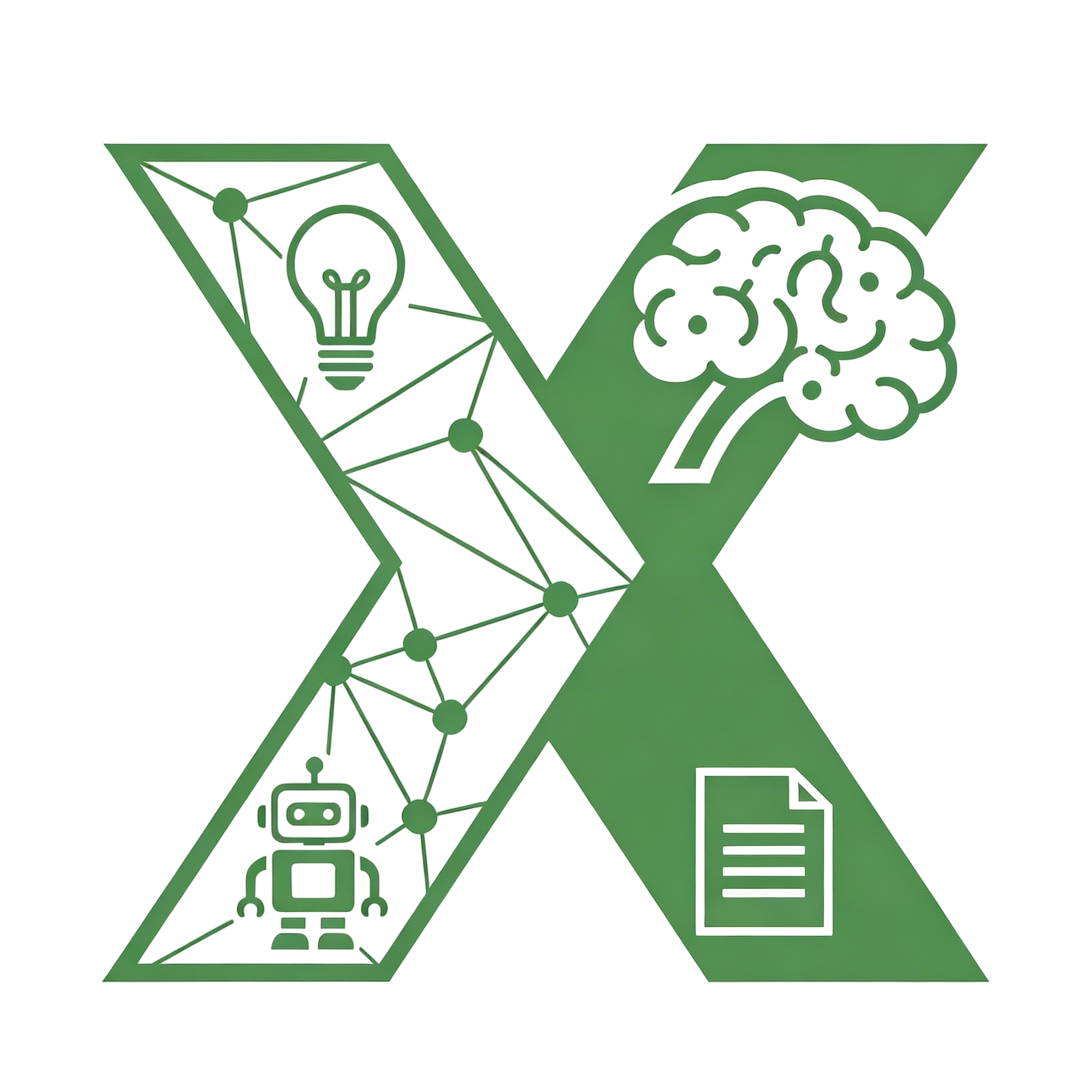}};
    \node[anchor=north east, at=(frame.north east), xshift=-0.5cm, yshift=-0.45cm] 
        {\includegraphics[height=1.00cm]{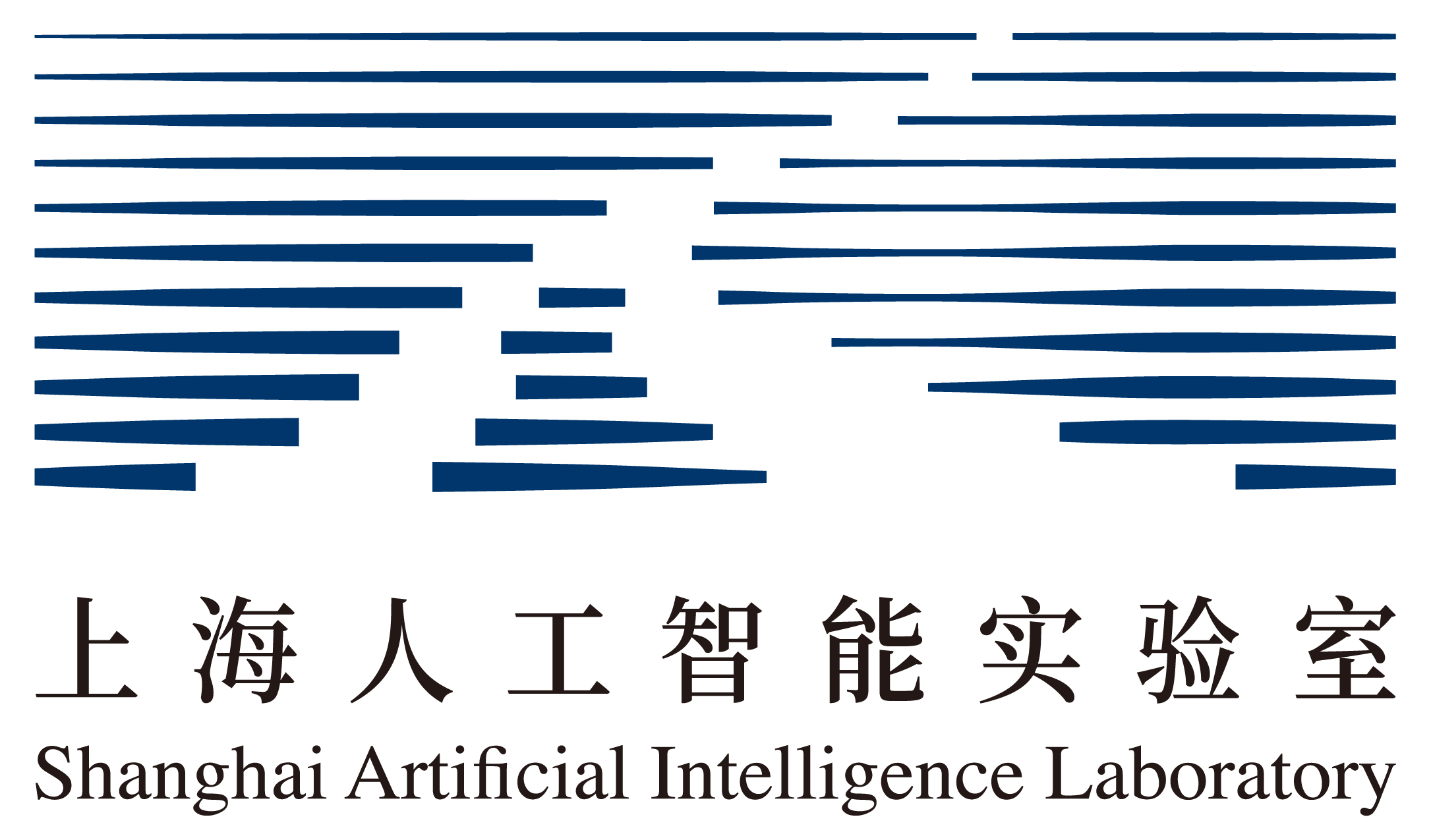}};
    }
  }%
  \begin{tcolorbox}
    \setlength{\parindent}{0cm}
    \setlength{\parskip}{0.5cm}
    {
      \setlength{\parskip}{0cm}
      \raggedright
      \nohyphens
      {
        \vskip 1.35cm 
        \setstretch{1.24}
        {\LARGE\sffamily\bfseries\textcolor{black}{\paperTitle}}\par
      }
      \vskip 0.25cm
      \paperAuthors\par
      \vskip 0.35cm
      \paperAffiliations\par
      \vskip 0.08cm
      \paperNotes\par
    }
    \vskip 0.2cm
    {\color{textgray}%
    \begin{abstract}

Reward-based reinforcement learning (RL) improves large language models but incurs substantial costs when repeated across targets. On-policy distillation (OPD) reuses experts by matching their distributions on target-generated trajectories, yet differing base policies can confound reward-induced updates with pre-existing policy differences, potentially distorting transferred update directions. We propose Proxy OPD (P-OPD), a post-training method that transfers reward-induced relative policy updates using the target's frozen base policy as a reference. A proxy undergoes reward-guided exploration, yielding token-level log-probability changes between its initial and optimized policies. These signals guide learning on target-generated trajectories, with calibration controlling signal absorption, enabling reuse of reward-induced improvements across targets. Experiments on Qwen3-family targets in mathematical reasoning, code generation, and instruction following show that P-OPD improves targets and outperforms conventional OPD in most settings. For 1.7B-to-8B transfer, average Math Mean@16 and Code Mean@8 scores exceed OPD by 8.3 and 3.4 points, respectively. Mechanism analyses show more faithful reward-induced supervision and controllable adaptation. Cross-scale and cross-generation experiments confirm signal transferability and reusability, while multi-domain distillation reveals synergistic gains and cross-domain interference. In the measured 4B-to-8B setting, target adaptation after proxy exploration takes approximately 15\% of direct target RL time; repeated signal reuse amortizes the one-time exploration cost across compatible models. These findings support relative policy updates as effective, transferable, and composable supervision for an ``explore once, transfer multiple times'' post-training paradigm.

\end{abstract}

\par}
    \vskip 0.2cm
    {
      \setlength{\parskip}{0cm}
      {\small {\sffamily\bfseries \raisebox{-0.2em}{\includegraphics[width=0.025\linewidth]{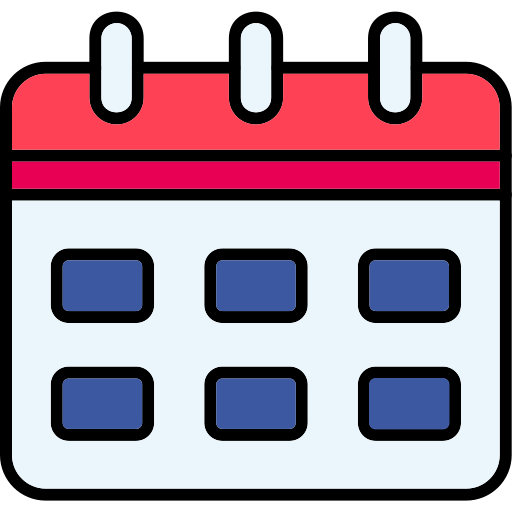}}~~Date:} \publishDate}\par%
      \vskip 0.08cm%
      {\small {\sffamily\bfseries \raisebox{-0.2em}{\includegraphics[width=0.025\linewidth]{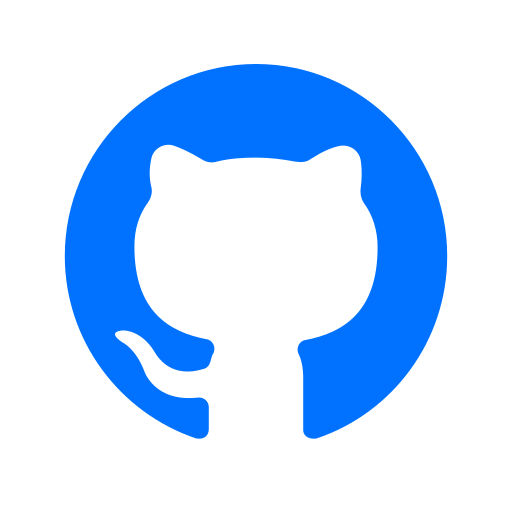}}~~Github Repo:} \githubLink}
      \vskip 0.08cm%
      {\small {\sffamily\bfseries \raisebox{-0.2em}{\includegraphics[width=0.025\linewidth]{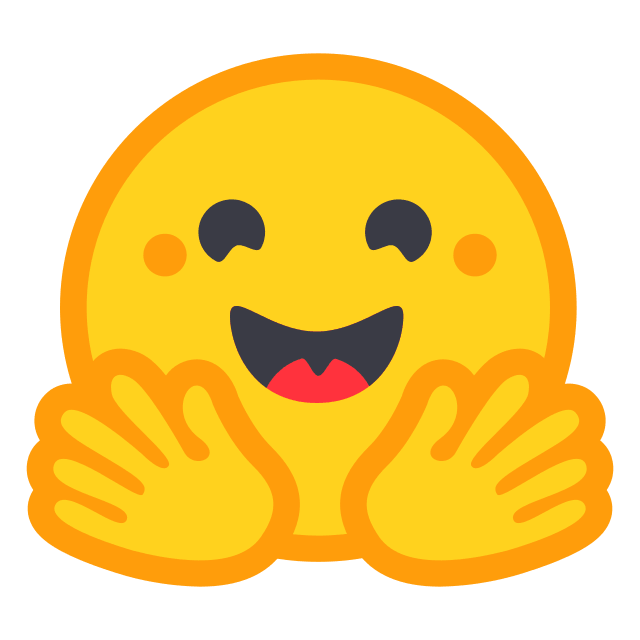}}~~Huggingface Page:} \huggingfaceLink}\par%
    }
  \end{tcolorbox}
  \tcbset{reset}
}

\newcommand{\methodFull}{\textbf{P}roxy \textbf{OPD}}
\newcommand{\methodAbb}{P-OPD}

\begin{document}

\newgeometry{top=1in, bottom=0.75in, textwidth=6.3in, textheight=9in}
\renderFrontBox



\section{Introduction}
\label{sec:intro}

\begin{figure*}
    \centering
    \includegraphics[width=0.98\linewidth]{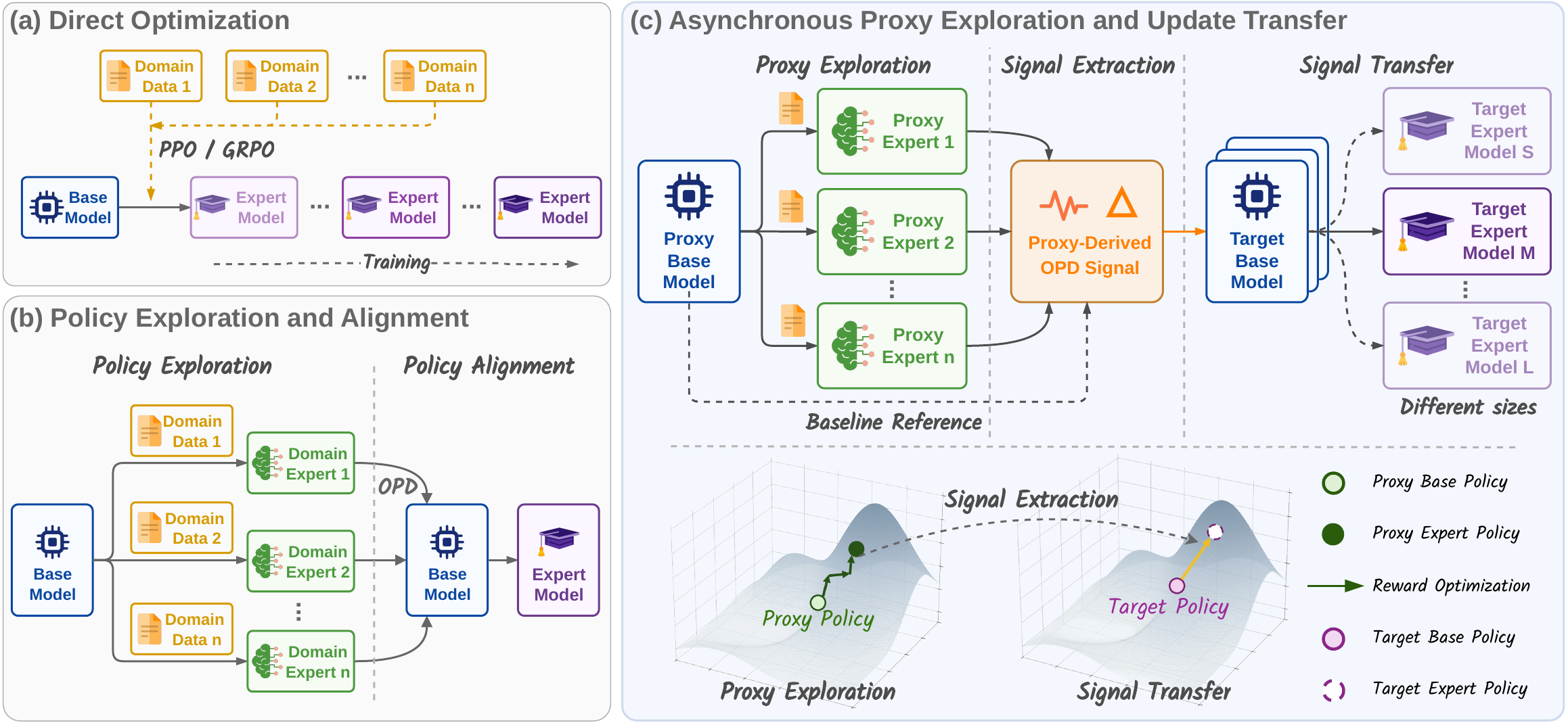}
    \caption{\textbf{Comparison of post-training pipelines.}
    \textbf{(a) Direct Optimization}: The target model is optimized directly using reward feedback.
    \textbf{(b) Policy Exploration and Alignment}: Policies are independently optimized into experts and then used to align the target model through distribution matching.
    \textbf{(c) Asynchronous Proxy Exploration and Update Transfer (Ours)}:
    Proxy policies are explored independently, with their relative on-policy updates extracted and transferred to the target model.}
    \label{fig:post_training_pipeline}
    \vspace{-0.5em}
\end{figure*}

Post-training plays a crucial role in enhancing the capabilities of large language models (LLMs)~\cite{lai2025survey,ouyang2022training}. Reward-based reinforcement learning (RL), such as PPO~\cite{schulman2017proximal} and GRPO~\cite{shao2024deepseekmath,guo2025deepseek}, improves model performance through iterative on-policy exploration and optimization guided by reward feedback. However, in direct RL, exploration is tightly coupled to the policy being optimized. Adapting a new target model typically requires collecting new trajectories and repeating reward-guided optimization, even when relevant capabilities have already been acquired by other models. As model scales, generations, and task domains continue to diversify, repeatedly performing such exploration introduces substantial computational overhead (Fig.~\ref{fig:post_training_pipeline}(a)). This raises a fundamental question: \emph{how can the policy improvements discovered through reward-guided exploration be effectively reused across different target models?}

On-policy distillation (OPD)~\cite{agarwal2024policy,song2026survey} provides a promising approach to addressing this challenge by decoupling \emph{expert exploration} from \emph{target-policy optimization}. Instead of performing reward optimization independently for each target policy, OPD leverages previously optimized expert policies to supervise different target models through distribution matching on target-generated trajectories~\cite{gu2024minillm}. This allows independently trained experts to be reused across targets (Fig.~\ref{fig:post_training_pipeline}(b)), avoiding repeated reward-guided exploration. However, conventional OPD directly aligns the target policy with the expert's absolute output distribution, without distinguishing the \emph{\textbf{improvements acquired through reward optimization}} from the expert's pre-existing preferences.

In fact, this distinction is critical for effective knowledge transfer across models. An expert's final distribution reflects both its initial policy preferences and the changes induced by reward optimization. When the expert and target have different base-policy distributions, directly matching the expert distribution inevitably entangles the desired reward-induced improvements with pre-existing model-specific differences. Such base-policy mismatch may introduce irrelevant supervision and drive the target policy away from its own preferences, rather than faithfully transferring the improvements discovered during exploration. For example, even if reward optimization leaves the expert unchanged, conventional OPD may still push the target toward the expert's original distribution. This reveals a fundamental limitation of absolute distribution matching: \emph{reusing an expert policy is not equivalent to reusing what the expert has learned}. Instead of transferring the expert's final distribution, which reflects what the expert prefers, we argue that a more appropriate form of reusable supervision is the \textbf{relative policy update signal}, which captures what the expert has learned through reward optimization~\cite{mitchell2023emulator, liu2024tuning}.

Motivated by this insight, we propose \methodFull{} (\textbf{\methodAbb{}}), a signal-centric post-training framework that transfers reward-induced relative policy updates rather than absolute expert distributions. \methodAbb{} consists of three stages: \textbf{proxy exploration}, \textbf{update-signal extraction}, and \textbf{signal transfer}. First, a proxy policy undergoes reward-guided optimization to discover task-specific policy improvements. Second, we extract the relative update signal as the token-level log-probability difference between the optimized proxy expert and its initial policy, isolating the policy changes induced by reward optimization. Finally, the extracted signals are transferred along target-generated trajectories, with the target's frozen initial policy serving as the reference. A calibration coefficient controls the strength of signal transfer, allowing the target to incorporate proxy-discovered improvements while limiting excessive deviations from its own base policy. By shifting the object of transfer from an expert's final distribution to its relative policy updates, \methodAbb{} enables the same exploration outcome to guide different target models while reducing interference from base-policy mismatch. Furthermore, proxy exploration and target optimization can proceed independently, supporting an asynchronous ``explore once, transfer multiple times'' post-training paradigm (Fig.~\ref{fig:post_training_pipeline}(c)).

Experiments on Qwen3-family models~\cite{yang2025qwen3} across mathematical reasoning, code generation, and instruction following. \methodAbb{} improves target models and outperforms conventional OPD in most settings. In 1.7B-to-8B transfer, average Math Mean@16 and Code Mean@8 scores exceed OPD by 8.3 and 3.4 points, respectively (Table~\ref{tab:math_code_performance}). Mechanism analyses show that relative updates provide more faithful reward-induced supervision, while calibration controls how strongly the target absorbs these signals. Cross-scale and cross-generation results demonstrate signal transferability and reusability, with an 8.4-point average Math Mean@16 gain over OPD when transferring from Skywork-OR1-Math-7B to Qwen3-8B (Table~\ref{tab:cross_gen_reusability}). Joint multi-domain transfer further demonstrates signal composability, yielding Math and Code gains alongside an instruction-following decline relative to single-domain transfer. Finally, after proxy exploration, 4B-to-8B adaptation takes approximately 15\% of direct 8B-RL400 training time; including one-time exploration, reusing a 1,200-step 4B proxy run across four compatible targets is projected to reduce cumulative training time by approximately 15.3\%.

Our main contributions are as follows:
\begin{itemize}[leftmargin=1em, itemsep=0pt, topsep=2pt]
    \item We reveal that conventional OPD entangles reward-induced policy improvements with model-specific preferences, motivating relative policy updates as a more faithful and reusable form of supervision.

    \item We propose \methodFull{}, a signal-centric post-training framework that extracts relative update signals from proxy base–expert pairs and transfers them on target-generated trajectories through target-anchored calibration, enabling independent proxy exploration and flexible signal reuse.

    \item We demonstrate that \methodAbb{} provides effective, transferable, and composable supervision through relative update signals across models and task domains, enabling flexible policy adaptation and efficient reuse of reward-guided exploration outcomes.
\end{itemize}
\section{Related Works}
\label{app:related_works}

We review reward-based policy optimization, on-policy distillation, and relative update transfer, highlighting the distinction between reusing an expert's final policy and reusing the improvements acquired through reward optimization.

\subsection{Reward-Based Policy Optimization}

Reward-based optimization improves policies through reward-guided exploration and learning. PPO~\cite{schulman2017proximal,christiano2017deep} provides a standard policy-optimization framework, while GRPO~\cite{shao2024deepseekmath,yu2026dapo} estimates advantages from groups of responses without a separate critic. Reward design also affects credit assignment: process supervision evaluates intermediate reasoning steps~\cite{lightman2024let,lee2023rlaif}, and token-level rewards provide finer-grained feedback for policy optimization~\cite{yoon2024tlcr}. These methods improve how policies discover and learn preferred behaviors, but typically require collecting and evaluating new rollouts for each target model. This motivates reusing previously discovered policy improvements across models rather than repeating reward-guided exploration for every target.

\subsection{On-Policy Distillation and Distribution Matching}

Knowledge distillation transfers supervision through teacher predictions~\cite{hinton2015distilling, wen2023f} or generated sequences~\cite{kim2016sequence,ko2024distillm}. Related preference-based methods, such as DPO~\cite{rafailov2023direct, ethayarajh2024model}, optimize policies from preference pairs~\cite{azar2024general,hong2024orpo}. On-policy distillation (OPD)~\cite{agarwal2024policy} instead queries teacher distributions on student-generated trajectories, aligning supervision with the student's own behavior~\cite{ross2011reduction,busbridge2025distillation}. G-OPD~\cite{yang2026learning} further studies the roles of reference-policy selection and reward scaling in this formulation. OPD enables expert reuse without repeating reward optimization for each target. However, matching absolute expert distributions entangles reward-induced improvements with pre-existing model-specific preferences. Our work shifts the transfer object from the expert's final distribution to its relative policy updates, aiming to reuse what the expert has learned through reward optimization and thereby provide more effective supervision.

\subsection{Relative Update Transfer and Signal Reuse}

Policy differences offer an alternative to absolute-distribution matching by capturing what a model has learned during optimization relative to its initial policy. Proxy-Tuning~\cite{liu2024tuning} adjusts target logits using the difference between tuned and untuned small-model logits at decoding time, without updating target parameters. WSPO~\cite{zhu2025weak} fits relative sequence-level log-probability changes on offline data. In contrast, \methodAbb{} transfers reward-induced token-level updates on target-generated trajectories, with the target's base policy as the reference and calibration controlling their influence. This provides target-adapted supervision that captures learned improvements rather than absolute expert distributions.

Concurrent work shares the principle of transferring relative policy changes rather than absolute expert distributions. Direct-OPD~\cite{feng2026weak} studies weak-to-strong transfer through a teacher's pre-/post-RL log-ratio with student-reference KL regularization, while OPD$^2$~\cite{heo2026onpolicydeltadistillation} combines a teacher–base delta signal with centering and OPD-direction gating for reasoning distillation. 
However, a systematic understanding of what makes relative signals effective across models and domains, how independently explored domain signals reinforce or interfere with one another, and when exploration reuse becomes economical remains incomplete. Developed independently, \methodAbb{} addresses these questions through mechanistic analyses, cross-model transfer and multi-domain composition experiments, and exploration-budget and cost analyses. We characterize the fidelity, transferability, and composability of relative updates, as well as the conditions under which repeated reuse lowers training costs. These findings clarify both the benefits and limitations of signal reuse, providing mechanistic and practical support for an ``explore once, transfer multiple times'' post-training paradigm.
\section{Methodology}
\label{sec:method}

\begin{figure*}
    \centering
    \includegraphics[width=\linewidth]{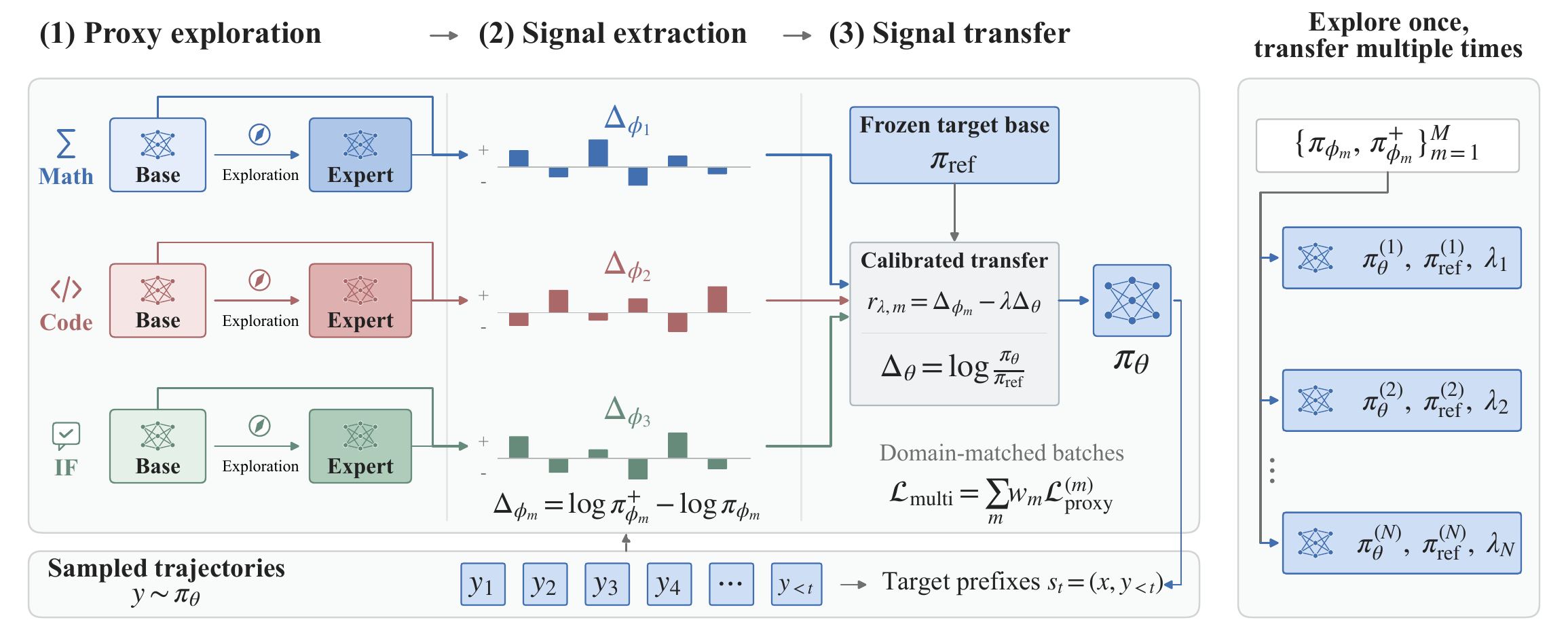}
    \caption{\textbf{Overview of \methodAbb{}.} Domain-specific proxies independently explore policy improvements, which are extracted as relative update signals and absorbed by the same target on its own trajectories with calibrated base-policy regularization. The explored proxy pairs can be reused across targets, enabling ``explore once, transfer multiple times.''}
    \label{fig:method}
\end{figure*}

Distribution matching provides an efficient way to absorb supervision from reward-optimized policies. A natural instantiation is to optimize a proxy model with rewards and then distill its output distribution into a target model. However, the proxy's absolute output distribution includes both its initial preferences and the changes discovered during exploration. When proxy and target bases differ, direct matching therefore transfers a model-dependent base-policy mismatch along with the useful update. To make exploration outcomes reusable across targets, we separate the relative update from the proxy's base distribution and transfer it with the target's own base policy as the reference. Our method consists of three stages: \emph{proxy exploration}, \emph{update-signal extraction}, and \emph{signal transfer}.

\subsection{Setup}

Let $x$ denote an input prompt and $y=(y_1,\ldots,y_T)$ denote a generated response. At decoding step $t$, we define the token-level state as
\begin{equation}
   s_t = (x, y_{<t}).
\end{equation}

Our framework involves two model families sharing a common vocabulary $\mathcal{V}$: a proxy family used for reward exploration, and a target family that we ultimately aim to improve. We use the following four policies throughout this section: (1) \textbf{Proxy base policy} $\pi_\phi$: the proxy policy before reward optimization; (2) \textbf{Proxy expert policy} $\pi_\phi^+$: the optimized proxy policy obtained after reward optimization; (3) \textbf{Target base policy} $\pi_{\mathrm{ref}}$: the frozen initial policy of the target model; (4) \textbf{Target policy} $\pi_\theta$: the trainable target policy being optimized. The proxy expert $\pi_\phi^+$ can be produced from $\pi_\phi$ during proxy exploration or supplied by an existing base--expert pair for reuse. In either case, $\pi_{\mathrm{ref}}$ remains fixed as the anchor for training $\pi_\theta$.

\subsection{OPD as a Direct Baseline}
\label{subsec:opd_baseline}

A straightforward way to use the proxy expert policy $\pi_\phi^+$ is on-policy distillation (OPD), which directly aligns the target policy $\pi_\theta$ to the proxy expert distribution:
\begin{equation}
    \mathcal{L}_{\mathrm{OPD}}(\theta)
    =
    \mathbb{E}_{x\sim\mathcal{D}}
    \left[
        D_{\mathrm{KL}}
        \left(
            \pi_\theta(\cdot\mid x)
            \,\|\,
            \pi_\phi^+(\cdot\mid x)
        \right)
    \right].
    \label{eq:opd}
\end{equation}
At the token level, this objective can be written as

\begin{equation}
    \mathcal{L}_{\mathrm{OPD}}(\theta)
    =
    -\mathbb{E}_{s_t\sim d_\theta}
    \left[
        \sum_{a\in\mathcal{V}}
        \pi_\theta(a\mid s_t)
        \log
        \frac{
            \pi_\phi^+(a\mid s_t)
        }{
            \pi_\theta(a\mid s_t)
        }
    \right].
    \label{eq:token_opd}
\end{equation}

Although simple, OPD directly uses the distribution of $\pi_\phi^+$ as the alignment target. Its token-level utility also reveals why this target can be mismatched. For a fixed state, the log-ratio implicit in Eq.~\ref{eq:token_opd} can be decomposed as

{\small
\begin{equation}
    \log \frac{\pi_\phi^+(a\mid s_t)}{\pi_\theta(a\mid s_t)}
    =
    \log \frac{\pi_\phi^+(a\mid s_t)}{\pi_\phi(a\mid s_t)}
    -
    \log \frac{\pi_\theta(a\mid s_t)}
              {\pi_{\mathrm{ref}}(a\mid s_t)}
    +
    \log \frac{\pi_\phi(a\mid s_t)}
              {\pi_{\mathrm{ref}}(a\mid s_t)}.
    \label{eq:opd_decomposition}
\end{equation}
}

The first term is the policy change induced by reward optimization, while the last term depends only on the mismatch between the two initial policies. Thus, direct OPD couples the desired reward-induced update with a model-dependent base-policy difference. When the proxy and target bases have different capabilities or priors, this additional term can provide a less reliable direction for improving the target policy. This motivates us to transfer the relative reward-induced update while retaining the target base as the reference policy.

\subsection{Stage 1: Proxy Exploration}
\label{subsec:proxy_explore}

The goal of proxy exploration is to obtain the proxy expert policy $\pi_\phi^+$ from the proxy base policy $\pi_\phi$. Direct reward optimization on the target model can be expensive and inflexible, especially when adapting to different reward functions. We therefore decouple reward exploration from target-model training by using the proxy model as an efficient exploration testbed.

Given a reward function $r(x,y)$, we optimize the proxy base policy $\pi_\phi$ using standard reward optimization algorithms, such as PPO or GRPO:
\begin{equation}
    \pi_\phi^+
    \leftarrow
    \operatorname{RewardOpt}
    \left(
        \pi_\phi,\,
        r
    \right).
    \label{eq:proxy_exploration}
\end{equation}
The resulting proxy expert policy $\pi_\phi^+$ captures reward-preferred behavioral changes discovered through trial-and-error exploration.

This stage avoids repeatedly running costly reward optimization on the target model. It also allows multiple proxy base policies to be optimized in parallel under different rewards or data distributions, producing diverse proxy expert policies whose update signals can be extracted and reused for subsequent transfer.

\subsection{Stage 2: Update-Signal Extraction}
\label{subsec:signal_extract}

After proxy exploration, standard distillation would directly use the proxy expert policy $\pi_\phi^+$ as the target. In contrast, we extract the \emph{relative} change from the proxy base policy $\pi_\phi$ to the proxy expert policy $\pi_\phi^+$.

At each prefix $s_t$ visited by the current target policy, both frozen proxy policies are evaluated on that same prefix. For each token $a\in\mathcal{V}$, we define the proxy-induced update signal as
\begin{equation}
        \Delta_\phi(a\mid s_t)
        =
        \log \pi_\phi^+(a\mid s_t)
        -
        \log \pi_\phi(a\mid s_t)
        =
        \log
        \frac{\pi_\phi^+(a\mid s_t)}
             {\pi_\phi(a\mid s_t)}.
    \label{eq:proxy_direction}
\end{equation}

Crucially, $\Delta_\phi$ is a log-ratio rather than an absolute probability distribution. It therefore represents the change caused by reward optimization, while removing the proxy base component from the OPD decomposition in Eq.~\ref{eq:opd_decomposition}. The signal can consequently be reused with a target model whose base distribution is different from the proxy base. We further introduce a scaling coefficient $\alpha>0$ and use $\alpha\Delta_\phi$ as the transferred signal, where larger $\alpha$ amplifies the proxy-discovered update and smaller $\alpha$ yields more conservative transfer.

\subsection{Stage 3: Signal Transfer}
\label{subsec:signal_transfer}

We now transfer the extracted proxy update signal to the target model. The proxy signal $\Delta_\phi$ indicates how reward optimization changes the proxy policy at each state-token pair. However, directly applying this signal to the target model may be overly aggressive, since the proxy and target models can have different base distributions and different capabilities. Therefore, signal transfer should both encourage the target policy to follow the proxy-induced update direction and regularize it against excessive deviation from its own base policy.

To this end, we optimize the target policy $\pi_\theta$ with two terms. The first term rewards tokens that are upweighted by proxy reward optimization, measured by $\Delta_\phi(a\mid s_t)$. The second term penalizes the target policy change relative to its frozen base policy $\pi_{\mathrm{ref}}$. Specifically, we define the target-policy change as
\begin{equation}
    \Delta_\theta(a\mid s_t)
    =
    \log
    \frac{
        \pi_\theta(a\mid s_t)
    }{
        \pi_{\mathrm{ref}}(a\mid s_t)
    }.
    \label{eq:primary_change}
\end{equation}
We then formulate the signal-transfer objective as
\begin{equation}
    \max_{\theta}
    \;
    \mathbb{E}_{s_t\sim d_\theta}
    \left[
        \sum_{a\in\mathcal{V}}
        \pi_\theta(a\mid s_t)
        \left(
            \alpha \Delta_\phi(a\mid s_t)
            -
            \beta \Delta_\theta(a\mid s_t)
        \right)
    \right],
    \label{eq:alpha_beta_objective}
\end{equation}
where $\alpha>0$ controls the strength of the transferred proxy update signal, and $\beta>0$ controls the conservatism of the target policy with respect to its own base policy. A larger $\alpha$ makes the target model more strongly absorb the proxy-discovered reward direction, while a larger $\beta$ discourages large deviations from $\pi_{\mathrm{ref}}$.

Since multiplying the objective by a positive constant preserves its optimizer, we divide Eq.~\ref{eq:alpha_beta_objective} by $\alpha$ and define
\begin{equation}
    \lambda = \frac{\beta}{\alpha}.
    \label{eq:lambda_def}
\end{equation}
This yields the calibrated token-level utility
\begin{equation}
    r_{\lambda}(a\mid s_t)
    =
    \Delta_\phi(a\mid s_t)
    -
    \lambda \Delta_\theta(a\mid s_t).
    \label{eq:calibrated_utility}
\end{equation}
Accordingly, the final signal-transfer loss is
\begingroup
{\small
\begin{equation}
    \begin{aligned}
        \mathcal{L}_{\mathrm{proxy}}(\theta) =
        -\mathbb{E}_{s_t\sim d_\theta}
        \Biggl[
        \sum_{a\in\mathcal{V}}
        \pi_\theta(a\mid s_t)
        \Bigl(
            \Delta_\phi(a\mid s_t)
            -
            \lambda\Delta_\theta(a\mid s_t)
        \Bigr)
        \Biggr].
    \end{aligned}
    \label{eq:proxy_loss}
\end{equation}
}
\endgroup

Under this normalized objective, $\lambda$ is the calibration coefficient governing the balance between signal absorption and base-policy regularization. It determines how conservatively the target policy absorbs the proxy-induced update signal. A larger $\lambda$ imposes a stronger penalty on deviations from the base policy and therefore leads to more conservative updates, whereas a smaller $\lambda$ allows the target policy to more aggressively follow the reward-induced direction discovered by the proxy.

For a fixed state and an unrestricted target distribution, the corresponding optimum has the closed form
\begin{equation}
    \pi^*(a\mid s_t)
    =
    \frac{\pi_{\mathrm{ref}}(a\mid s_t)
    \exp\!\left(\Delta_\phi(a\mid s_t)/\lambda\right)}
    {\sum_{a'\in\mathcal{V}}\pi_{\mathrm{ref}}(a'\mid s_t)
    \exp\!\left(\Delta_\phi(a'\mid s_t)/\lambda\right)}.
    \label{eq:closed_form_target}
\end{equation}
This expression makes the role of calibration explicit: $\lambda$ acts as an update temperature, with $1/\lambda$ scaling the proxy direction: larger temperatures favor conservative behavior close to the target base, while smaller temperatures strengthen signal absorption. In practice, the target policy is parameterized by a neural network and optimized on sampled states, so Eq.~\ref{eq:closed_form_target} is an interpretation of the fixed-state objective rather than an explicitly constructed distribution.

During signal transfer, the proxy base policy $\pi_\phi$, the proxy expert policy $\pi_\phi^+$, and the target base policy $\pi_{\mathrm{ref}}$ remain frozen. Gradients are applied only to the target policy $\pi_\theta$.

\subsection{Multi-Domain Distillation}
\label{subsec:multi_domain}

The same formulation naturally supports multiple domains and multiple proxy experts. Let $m\in\{1,\ldots,M\}$ index domains, and let $\pi_{\phi_m}$ and $\pi_{\phi_m}^{+}$ denote the proxy base and proxy expert for domain $d$, respectively. Each proxy is optimized independently with its domain-specific reward or high-quality data distribution, producing the update signal
\begin{equation}
    \Delta_{\phi_m}(a\mid s_t)
    =
    \log \frac{\pi_{\phi_m}^{+}(a\mid s_t)}{\pi_{\phi_m}(a\mid s_t)}.
    \label{eq:domain_proxy_direction}
\end{equation}
All signals are transferred to the same target policy $\pi_\theta$, whose frozen target base $\pi_{\mathrm{ref}}$ is shared across domains. With domain sampling weights $w_m$ satisfying $w_m\geq 0$ and $\sum_m w_m=1$, multi-domain distillation minimizes
\begin{equation}
    \mathcal{L}_{\mathrm{multi}}(\theta)
    =
    \sum_{m=1}^{M} w_m
    \mathcal{L}_{\mathrm{proxy}}^{(m)}(\theta),
    \label{eq:multi_domain_loss}
\end{equation}
where $\mathcal{L}_{\mathrm{proxy}}^{(m)}$ is Eq.~\ref{eq:proxy_loss} with $\Delta_\phi$ replaced by $\Delta_{\phi_m}$ and the expectation taken over domain-$m$ states. In implementation, each batch contains data from one domain and uses only its corresponding proxy expert~\cite{deepseekai2026deepseekv4highlyefficientmilliontoken}; alternating batches optimize one shared target policy. This design allows domain experts to explore in parallel or asynchronously, while their relative update signals are jointly absorbed by a single target model without requiring the proxy experts themselves to be merged.

Overall, our method preserves the online distribution-matching nature of OPD, but replaces direct imitation of a proxy expert distribution with relative update-signal transfer. A single proxy can provide a reusable update direction, while multiple domain-specific proxies can jointly guide one target policy. The proxy models discover how reward optimization changes their policies, and the target policy absorbs these changes in its own base-policy-centered space.

\section{Experiments}

\begin{table*}[t]
\centering
\renewcommand{\arraystretch}{1.15}

\providecommand{\gain}[1]{\scriptsize{\textcolor{gray}{(+#1)}}}
\providecommand{\bestgain}[1]{\scriptsize{\textcolor{teal}{\textbf{(+#1)}}}}

\caption{
\textbf{Mean results of math, code, and instruction-following benchmarks}.
Absolute scores are followed by gains ($\Delta$) over the corresponding base models in parentheses. Gains are computed from the displayed scores: proxy experts are compared with their own proxy bases, and transferred targets with the target base.
The best absolute scores are highlighted in \textbf{bold}; \textcolor{teal}{\textbf{bold+teal}}
indicates the best gains.
In \texttt{OPD} and \texttt{P-OPD}, ``$\rightarrow$'' identifies the proxy expert on the left and the target policy on the right.
}

\resizebox{\textwidth}{!}{%
\begin{tabular}{l ccccc cccc c}
\toprule
\multirow{2}{*}{\raisebox{-1ex}{\textbf{Model}}}
& \multicolumn{5}{c}{\textbf{Math}}
& \multicolumn{4}{c}{\textbf{Code}}
& \multicolumn{1}{c}{\textbf{IF}} \\
\cmidrule(lr){2-6}
\cmidrule(lr){7-10}
\cmidrule(lr){11-11}
& \textbf{AIME24}
& \textbf{AIME25}
& \makecell[c]{\textbf{HMMT25}\\[-1pt]\textbf{(FEB)}}
& \makecell[c]{\textbf{HMMT25}\\[-1pt]\textbf{(NOV)}}
& \textbf{AVG.}
& \textbf{TACO}
& \makecell[c]{\textbf{Code}\\[-1pt]\textbf{Forces}}
& \textbf{APPs}
& \textbf{AVG.}
& \textbf{Nemotron} \\
\midrule

\multicolumn{11}{c}{\cellcolor{gray!15}\textit{Proxy Exploration}} \\

Qwen3-1.7B (Proxy Base)
& 11.9
& 11.0
& 5.8
& 5.2
& 8.5
& 7.3
& 17.1
& 12.0
& 12.1
& 21.7 \\

Qwen3-1.7B (Proxy Expert)
& 35.6 \gain{23.7}
& 31.7 \gain{20.7}
& 19.6 \gain{13.8}
& 15.2 \gain{10.0}
& 25.5 \gain{17.0}
& 19.1 \bestgain{11.8}
& 29.3 \gain{12.2}
& 38.7 \bestgain{26.8}
& 29.0 \bestgain{16.9}
& 46.3 \gain{24.6} \\

Qwen3-4B (Proxy Base)
& 23.2
& 22.1
& 10.6
& 7.9
& 16.0
& 12.6
& 20.9
& 29.4
& 21.0
& 29.4 \\

Qwen3-4B (Proxy Expert)
& 58.8 \gain{35.6}
& \textbf{55.8} \bestgain{33.7}
& 33.1 \bestgain{22.5}
& 38.3 \bestgain{30.4}
& 46.5 \bestgain{30.5}
& 19.6 \gain{7.1}
& 31.5 \gain{10.6}
& 38.6 \gain{9.2}
& 29.9 \gain{8.9}
& 56.1 \bestgain{26.8} \\

\midrule

\multicolumn{11}{c}{\cellcolor{gray!15}\textit{Update Signal Transfer}} \\

Qwen3-8B (Target Base)
& 24.4
& 22.5
& 12.3
& 10.0
& 17.3
& 14.3
& 27.2
& 30.8
& 24.1
& 30.6 \\

OPD (Proxy Expert 1.7B $\rightarrow$ Target Policy 8B)
& 41.0 \gain{16.6}
& 34.4 \gain{11.9}
& 19.2 \gain{6.9}
& 18.5 \gain{8.5}
& 28.3 \gain{11.0}
& 23.0 \gain{8.6}
& 36.8 \gain{9.6}
& 47.5 \gain{16.7}
& 35.8 \gain{11.6}
& 47.7 \gain{17.2} \\

\rowcolor{tabcolor!25}
\methodAbb{} (Proxy Expert 1.7B $\rightarrow$ Target Policy 8B)
& 50.8 \gain{26.4}
& 42.1 \gain{19.6}
& 22.9 \gain{10.6}
& 30.6 \gain{20.6}
& 36.6 \gain{19.3}
& \textbf{26.0} \gain{11.7}
& \textbf{40.2} \bestgain{13.0}
& \textbf{51.2} \gain{20.4}
& \textbf{39.2} \gain{15.0}
& 45.1 \gain{14.5} \\

OPD (Proxy Expert 4B $\rightarrow$ Target Policy 8B)
& 59.6 \gain{35.2}
& 50.6 \gain{28.1}
& 27.5 \gain{15.2}
& 36.7 \gain{26.7}
& 43.6 \gain{26.3}
& 19.6 \gain{5.3}
& 34.2 \gain{6.9}
& 41.9 \gain{11.1}
& 31.9 \gain{7.8}
& 54.3 \gain{23.8} \\

\rowcolor{tabcolor!25}
\methodAbb{} (Proxy Expert 4B $\rightarrow$ Target Policy 8B)
& \textbf{62.5} \bestgain{38.1}
& 52.7 \gain{30.2}
& \textbf{34.2} \gain{21.9}
& \textbf{40.4} \bestgain{30.4}
& \textbf{47.5} \gain{30.2}
& 21.8 \gain{7.5}
& 39.2 \gain{12.0}
& 44.7 \gain{13.9}
& 35.3 \gain{11.1}
& \textbf{56.3} \gain{25.7} \\

\bottomrule
\end{tabular}%
}


\label{tab:math_code_performance}
\end{table*}

In this section, we conduct comprehensive experiments to answer the following research questions:
\begin{itemize}[leftmargin=1em, itemsep=0pt, topsep=2pt]
    \item \textbf{Effectiveness \& Fidelity (Section~\ref{sec:effectiveness}):} Can \methodAbb{} provide more effective and faithful policy-improvement supervision than OPD across domains, and how does calibration regulate signal absorption?
    \item \textbf{Transferability (Section~\ref{sec:transferability}):} Do proxy-explored update signals remain effective across model scales and generations, as well as through sequential transfer?
    \item \textbf{Composability (Section~\ref{sec:exp_multi_domain}):} Can independently explored domain-specific update signals be jointly distilled into a single target model, and what synergy or interference arises from their combination?
    \item \textbf{Efficiency (Section~\ref{exp:efficiency}):} How do proxy exploration budget and signal reuse govern the trade-off between signal quality and total training cost across multiple targets?
\end{itemize}

\subsection{Effectiveness \& Fidelity}
\label{sec:effectiveness}

\subsubsection{Single-Domain Transfer}

To evaluate the effectiveness of \methodAbb{}, we conduct single-domain update-signal transfer experiments in three domains: Math, Code, and instruction following (IF). We use Qwen3-1.7B and Qwen3-4B as proxy base models, train domain-specific experts via GRPO, and transfer the resulting update signals to Qwen3-8B.
For Math, both proxy models undergo 500 steps of GRPO exploration on DeepMath-103K~\cite{he2025deepmath}\footnote{We filter the dataset using the selection strategy proposed in prior work~\cite{yang2026learning}.}.
For Code, we use Eurus-RL-Code~\cite{cui2025process}, with 300 exploration steps for Qwen3-4B and 500 for Qwen3-1.7B.
For IF, we split Nemotron~\cite{nvidia_nemotron_3_ultra_2026} into training and test sets comprising 80\% and 20\% of the data, respectively, and run 500 steps of GRPO exploration for both proxy models.
We select the calibration coefficient $\lambda$ based solely on preliminary experiments in the Math domain, setting $\lambda=1.0$ for 4B$\rightarrow$8B transfer and $\lambda=1.5$ for 1.7B$\rightarrow$8B transfer.
For each proxy-to-target configuration, the selected value is held fixed across all three domains and reused for Code and IF without additional tuning.
We evaluate Math performance on AIME24, AIME25, HMMT25 (Feb), and HMMT (Nov), reporting Mean@16.
For Code, we evaluate on TACO~\cite{li2023taco}, Codeforces~\cite{quan2025codeelobenchmarkingcompetitionlevelcode}, and APPS~\cite{hendrycksapps2021}, reporting Mean@8.
For IF, we evaluate on the held-out Nemotron test set, also reporting Mean@8.

As shown in Table~\ref{tab:math_code_performance}, \methodAbb{} improves Qwen3-8B over its base model on every evaluated benchmark with both proxy scales. Compared with OPD, \methodAbb{} achieves higher scores on all Math and Code benchmarks: the Math average increases from 28.3 to 36.6 with the 1.7B proxy and from 43.6 to 47.5 with the 4B proxy, while the Code average increases from 35.8 to 39.2 and from 31.9 to 35.3, respectively. On IF, the 4B proxy improves over OPD from 54.3 to 56.3, whereas the 1.7B proxy scores 45.1 versus OPD's 47.7; both exceed the target base score of 30.6. The larger proxy yields stronger Math and IF results, while the 1.7B proxy achieves the best Code average, showing that transfer quality depends on both proxy scale and domain.

\subsubsection{Reward Direction Fidelity}







\begin{wraptable}{r}{0.50\textwidth}
\vspace{-1.1em}
\centering
\renewcommand{\arraystretch}{1.25}
\setlength{\tabcolsep}{5pt}
\caption{\textbf{Performance under different $\lambda$ values.}
Mean@16 results on AIME24 and AIME25. Qwen3-1.7B Math (RL) 500 Steps is used as proxy expert for both OPD and Proxy-OPD}
\label{tab:ablation_study}
\resizebox{\linewidth}{!}{%
\begin{tabular}{@{}lcccccc@{}}
\toprule
\multirow{2}{*}{\textbf{Method}}
& \multicolumn{3}{c}{\textbf{AIME24}}
& \multicolumn{3}{c}{\textbf{AIME25}} \\
\cmidrule(lr){2-4}
\cmidrule(l){5-7}
& \multicolumn{1}{c}{$\lambda=1.25$}
& \multicolumn{1}{c}{$\lambda=1.50$}
& \multicolumn{1}{c}{$\lambda=1.75$}
& \multicolumn{1}{c}{$\lambda=1.25$}
& \multicolumn{1}{c}{$\lambda=1.50$}
& \multicolumn{1}{c}{$\lambda=1.75$} \\
\midrule
OPD Reward
& 45.3
& 46.2
& 46.8
& 37.2
& 39.4
& 37.9 \\
POPD Reward
& 50.8
& 50.8
& 51.7
& 37.9
& 42.1
& 39.2 \\
\bottomrule
\end{tabular}%
}
\vspace{-1em}
\end{wraptable}

To understand why \methodAbb{} outperforms OPD, we examine the fidelity of their implicit reward directions. Since the coefficient $\lambda$ controls how conservatively the target policy absorbs the transferred signal, a natural question is whether the performance advantage of \methodAbb{} arises from calibration rather than the reward signal itself. To disentangle these effects, we apply the same target-base regularization coefficient $\lambda$ to both OPD and \methodAbb{}, using Qwen3-1.7B Math (RL), trained for 500 steps, as the shared proxy expert. As shown in Table~\ref{tab:ablation_study}, varying $\lambda$ affects the performance of both methods, yet \methodAbb{} consistently achieves higher Mean@16 scores on both AIME24 and AIME25 across all tested values. This advantage under matched calibration settings motivates us to examine the quality of the underlying reward directions.

Eq.~\ref{eq:opd_decomposition} provides a potential explanation for this difference. Before target-policy regularization, the implicit reward in OPD can be written as
\begin{equation}
    r_{\mathrm{OPD}}(a\mid s_t)
    =
    \underbrace{
        \log \frac{\pi_\phi^+(a\mid s_t)}
                  {\pi_\phi(a\mid s_t)}
    }_{\text{proxy reward-induced update}}
    +
    \underbrace{
        \log \frac{\pi_\phi(a\mid s_t)}
                  {\pi_{\mathrm{ref}}(a\mid s_t)}
    }_{\text{base-policy mismatch}}.
\end{equation}
The first term is precisely the implicit reward transferred by \methodAbb{}, whereas the second reflects differences between the proxy and target base policies rather than changes induced by reward optimization. Consequently, OPD mixes the desired update signal with a model-dependent bias, which may distort the reward direction for the target model. We therefore investigate whether removing this bias yields a signal that better aligns with the target model's own reward-induced update.


\begin{figure}[t]
\centering
\begin{subfigure}[t]{0.44\linewidth}
    \centering
    \includegraphics[width=\linewidth]
        {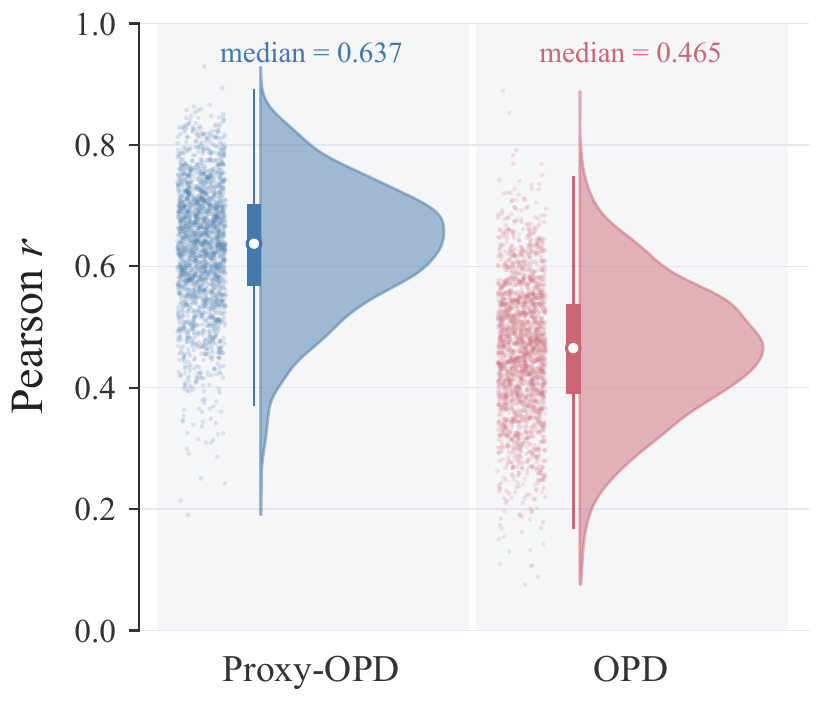}
    \caption{Pearson correlation.}
    \label{fig:reward-correlation-pearson}
\end{subfigure}%
\hspace{0.04\linewidth}%
\begin{subfigure}[t]{0.44\linewidth}
    \centering
    \includegraphics[width=\linewidth]
        {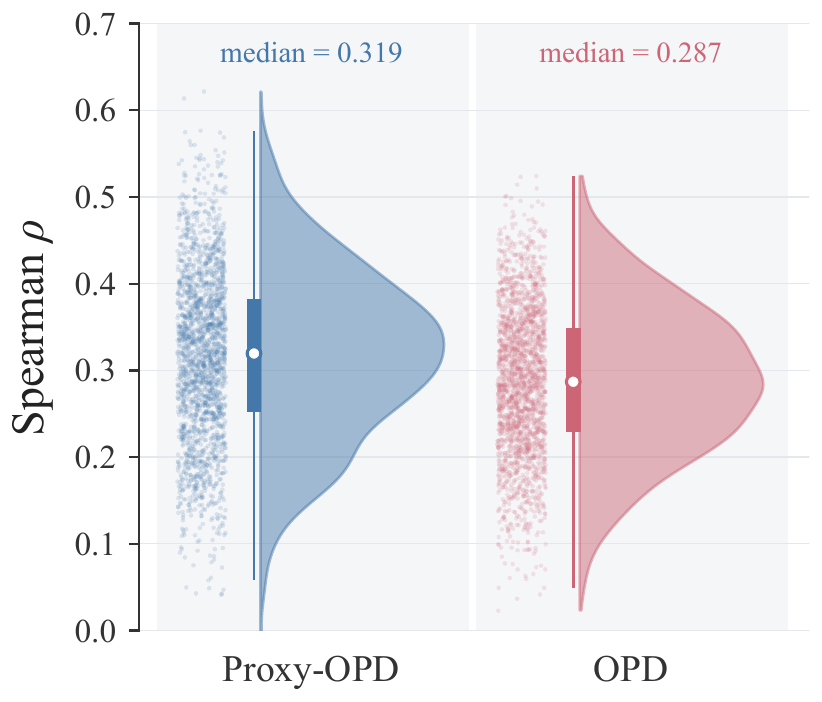}
    \caption{Spearman correlation.}
    \label{fig:reward-correlation-spearman}
\end{subfigure}
\caption{\textbf{Alignment of token-level implicit rewards with the target model's reward-induced update.}
The offline reference is the log-probability difference between the Qwen3-8B target expert after 400 GRPO steps and its base policy.
Each point represents a correlation computed across response tokens within one rollout, and the white marker denotes the median.}
\label{fig:reward-correlation}
\end{figure}

To construct an empirical ground-truth reference, we perform reward optimization directly on the target model. Importantly, GRPO uses sequence-level rewards and does not directly observe token-level rewards. A single optimization step therefore provides an imperfect estimate of the underlying token-level reward direction, owing to noisy exploration and indirect credit assignment. Rather than using a single-step policy change, we train the Qwen3-8B target base with GRPO on the Math training dataset for 400 steps to obtain a target expert. The cumulative log-probability change from the target base to this expert captures the direction learned through repeated reward-guided exploration and serves as a more stable offline reference. We treat this quantity as an empirical reference for the target model's implicit reward, rather than an exact token-level ground truth.

We then collect 2,048 rollouts from the frozen Qwen3-8B target base and evaluate all reward signals on the same response tokens and prefixes. Specifically, the reference reward is $\log p_{\mathrm{8B\text{-}RL400}}-\log p_{\mathrm{8B\text{-}base}}$, the \methodAbb{} reward is $\log p_{\mathrm{1.7B\text{-}expert}}-\log p_{\mathrm{1.7B\text{-}base}}$, and the OPD reward is $\log p_{\mathrm{1.7B\text{-}expert}}-\log p_{\mathrm{8B\text{-}base}}$. This comparison isolates the implicit reward components from target-policy regularization. For each rollout, we compute Pearson and Spearman correlations between each transferred reward and the reference reward across response tokens, yielding one coefficient per rollout for each method and metric.

As shown in Figure~\ref{fig:reward-correlation}, \methodAbb{} aligns more closely with the reference reward than OPD under both metrics. The median Pearson correlation is 0.637 for \methodAbb{} versus 0.465 for OPD, while the median Spearman correlation is 0.319 versus 0.287. Across all 2,048 rollouts, the corresponding mean correlations are 0.631 versus 0.462 for Pearson and 0.317 versus 0.287 for Spearman. These results are consistent with the decomposition in Eq.~\ref{eq:opd_decomposition}: measuring the proxy expert relative to its own base isolates the reward-induced update and avoids introducing the proxy--target base-policy mismatch. The resulting improvement in reward-direction fidelity offers a mechanistic explanation for the stronger downstream performance of \methodAbb{}.

\subsubsection{Role of the Calibration Term}

\begin{wrapfigure}{r}{0.48\linewidth}
\vspace{-1.2em}
\centering
\includegraphics[width=\linewidth]{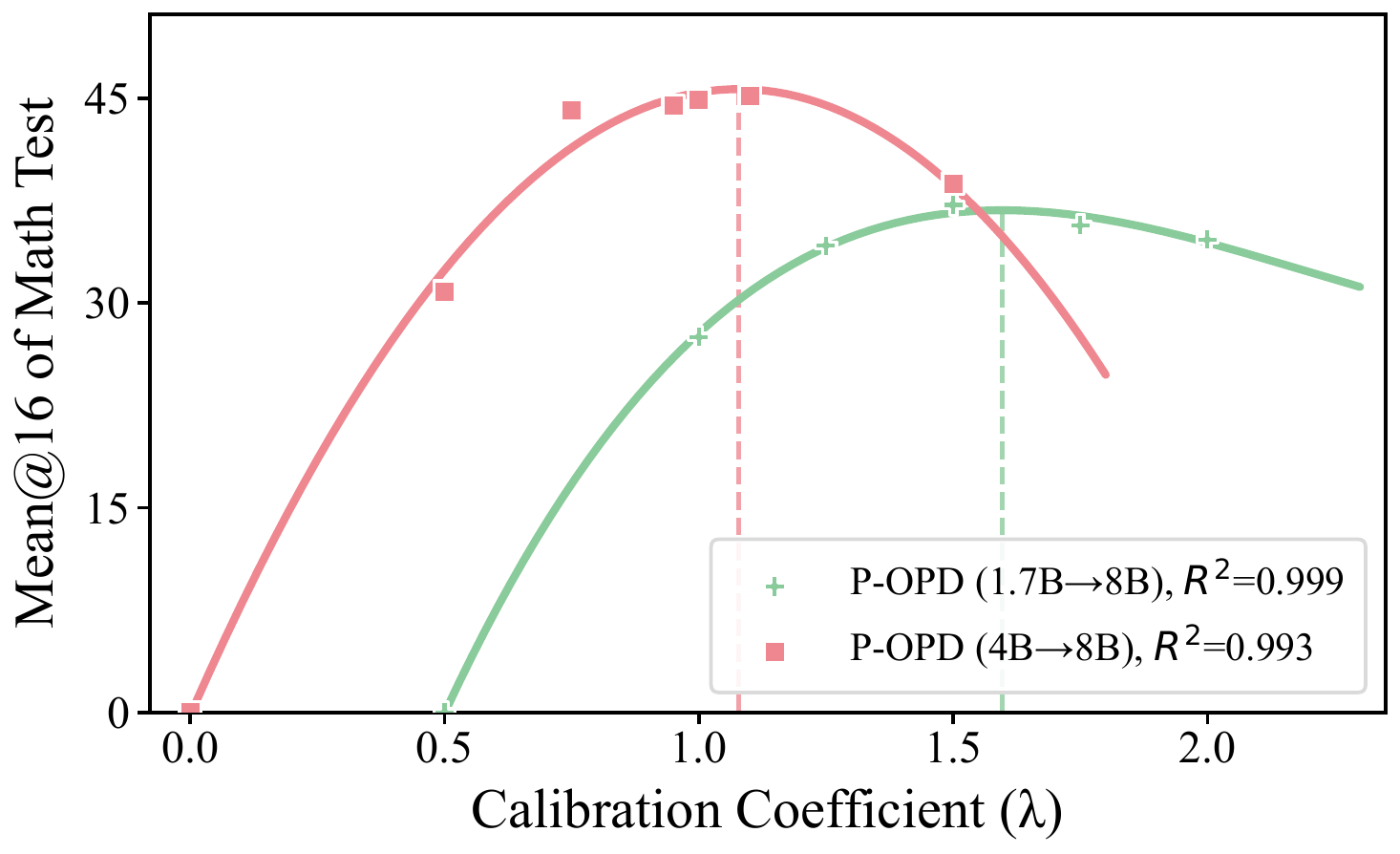}
\caption{\textbf{Sensitivity to proxy model scale and calibration coefficient.}
Each curve is obtained by fitting a cubic polynomial to the experimental results.
An $R^2$ value closer to 1 indicates a better fit.}
\label{fig:sensitivity_analysis}
\end{wrapfigure}

The calibration gains observed for both OPD and \methodAbb{} in Table~\ref{tab:ablation_study} motivate a closer examination of how this term regulates update-signal transfer. We first sweep $\lambda$ for Math transfer from Qwen3-4B and Qwen3-1.7B to Qwen3-8B, reporting average Mean@16 over eight Math benchmarks\footnote{Per-benchmark results are provided in Appendix~\ref{app:lambda_ablation}.}. As shown in Figure~\ref{fig:sensitivity_analysis}, cubic fits to the measured scores peak near $\lambda=1.08$ for the 4B proxy and $\lambda=1.57$ for the 1.7B proxy. The 4B proxy reaches a higher peak score, while the smaller proxy benefits from stronger calibration.

To explain this sensitivity, we examine how calibration shapes the target policy's distribution during transfer. The calibration term in Eq.~\ref{eq:proxy_loss} is a KL penalty relative to the frozen target base, making distribution migration a diagnostic. Since \methodAbb{} transfers relative updates rather than matching the proxy expert's absolute distribution, we define a \emph{virtual teacher} by applying the proxy update to the target base:
\begin{equation}
    q(a\mid s_t)
    =
    \frac{\pi_{\mathrm{ref}}(a\mid s_t)
    \exp\!\left(\Delta_\phi(a\mid s_t)\right)}
    {\sum_{a'\in\mathcal{V}}\pi_{\mathrm{ref}}(a'\mid s_t)
    \exp\!\left(\Delta_\phi(a'\mid s_t)\right)}.
    \label{eq:virtual_teacher}
\end{equation}
This distribution corresponds to Eq.~\ref{eq:closed_form_target} at $\lambda=1$ and provides a common reference across calibration settings. Both the virtual teacher and target base remain fixed for any given prefix because the proxy base--expert pair and target base are frozen.

We track their overlap with the target policy during 1.7B$\rightarrow$8B Math transfer with $\lambda\in\{1.25,1.5,1.75\}$. At each response position $t$, we form a candidate set $C_t$ as the union of the top-$K$ tokens from the proxy expert, current target policy, and target base. We renormalize each distribution over $C_t$ and measure probability-mass overlap as
{\small
\begin{equation}
    O_t(p,q)
    = \sum_{a\in C_t}\min\!\left(p_t^C(a),q_t^C(a)\right)
    = 1-\frac{1}{2}\sum_{a\in C_t}\left|p_t^C(a)-q_t^C(a)\right|.
    \label{eq:mass_overlap}
\end{equation}
}
Larger overlap means closer distributions on the candidate set. Since most positions show little disagreement, at each checkpoint we select the 10\% with the lowest virtual-teacher--target-base overlap and average all three pairwise overlaps over those same positions. This prevents largely agreeing positions from masking the migration dynamics. These averages also depend on the prefixes sampled by the target policy: visiting more mutually agreeable contexts can increase overlap even when the two underlying policies are frozen.

\begin{figure*}[t]
    \centering
    \begin{subfigure}[t]{0.32\textwidth}
        \centering
        \includegraphics[width=\linewidth]{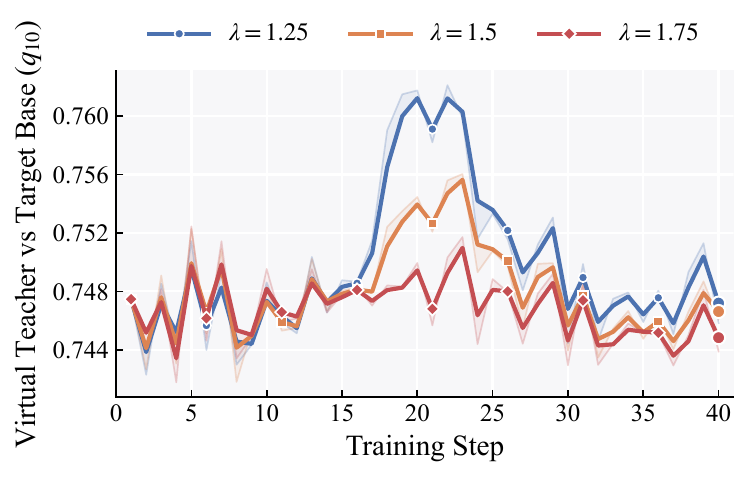}
        \caption{Virtual teacher vs. target base.}
        \label{fig:migration-virtual-base}
    \end{subfigure}
    \hfill
    \begin{subfigure}[t]{0.32\textwidth}
        \centering
        \includegraphics[width=\linewidth]{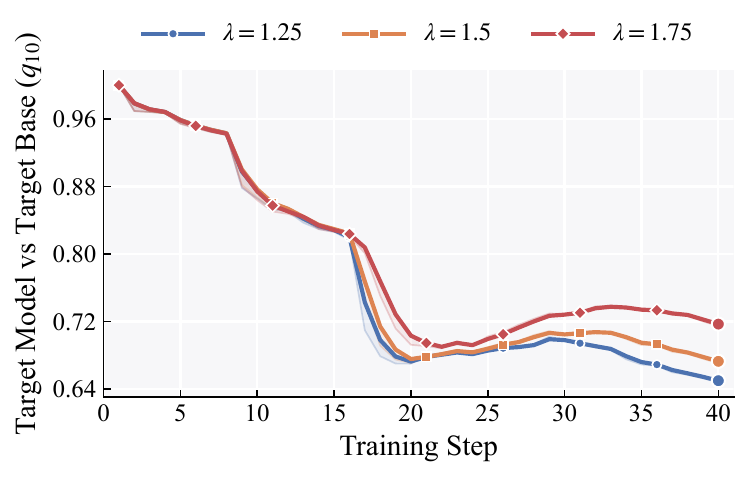}
        \caption{Target policy vs. target base.}
        \label{fig:migration-target-base}
    \end{subfigure}
    \hfill
    \begin{subfigure}[t]{0.32\textwidth}
        \centering
        \includegraphics[width=\linewidth]{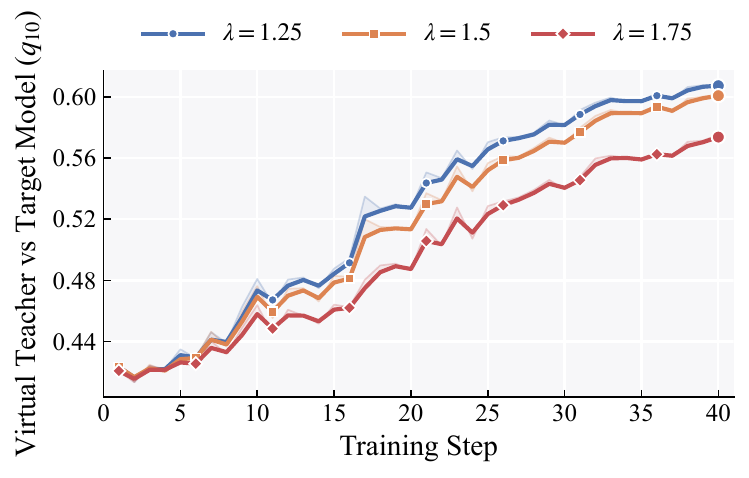}
        \caption{Virtual teacher vs. target policy.}
        \label{fig:migration-virtual-target}
    \end{subfigure}
    \caption{\textbf{Mass overlap ratio under different calibration coefficients.}
    All three panels use the same token positions, selected as the lowest 10\% by virtual-teacher--target-base mass overlap, to highlight positions with substantial distributional disagreement.}
    \label{fig:distribution-migration}
\end{figure*}

\begin{figure*}[t]
    \centering
    \includegraphics[width=0.98\linewidth]{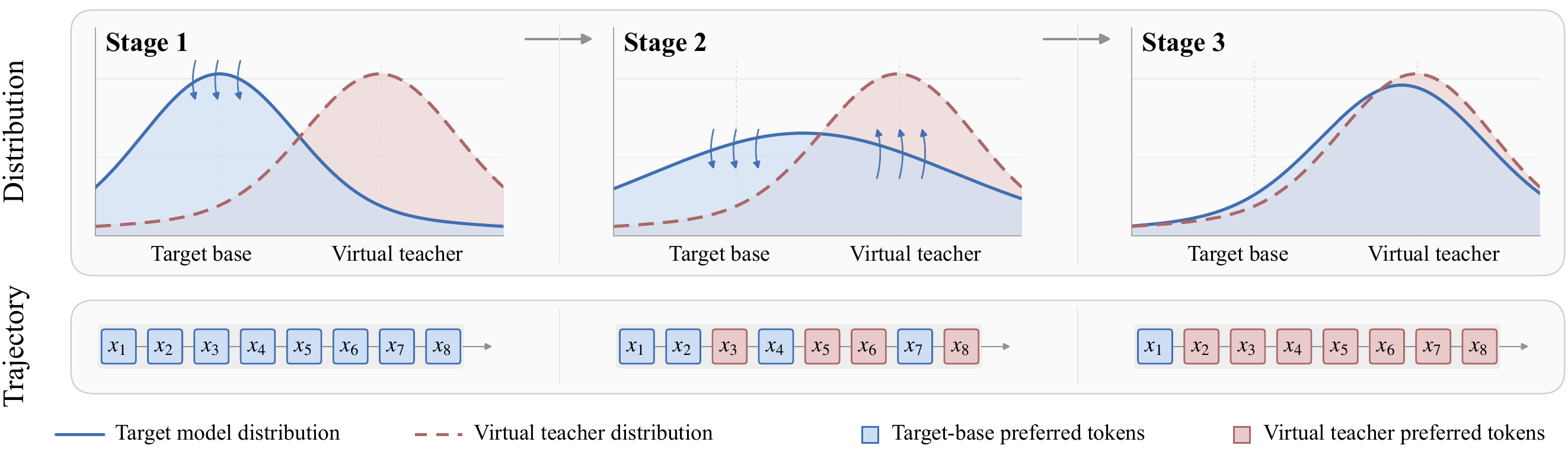}
    \vspace{-0.5em}
    \caption{\textbf{Joint migration of token distributions and sampled trajectories during update-signal transfer.}
    \textbf{Stage 1:} The target starts from its base policy and mainly samples base-preferred tokens; transfer suppresses dominant probabilities that conflict with the virtual teacher.
    \textbf{Stage 2:} The target distribution becomes flatter, allowing trajectories to include more teacher-preferred tokens alongside base-preferred tokens.
    \textbf{Stage 3:} Probability mass shifts toward teacher-preferred tokens, and sampled trajectories increasingly reflect the virtual teacher's preferences.}
    \label{fig:ill_distribution_migration}
    \vspace{-1em}
\end{figure*}

Figure~\ref{fig:distribution-migration} shows that virtual-teacher--target-base overlap initially rises and then falls across the tested coefficients. Meanwhile, target-policy--base overlap declines overall, and virtual-teacher--target-policy overlap increases. Since the virtual teacher and target base are frozen, changes in their measured overlap reflect changes in the visited prefixes and candidate sets as the target's on-policy trajectories evolve. The curves thus capture migration of both the target distribution and the sampled trajectories.

Figure~\ref{fig:ill_distribution_migration} illustrates this progression in three stages. Initially, the target policy samples mainly tokens favored by its base, and transfer suppresses dominant probabilities that conflict with the virtual teacher. The resulting flatter distribution makes teacher-preferred tokens easier to sample, producing trajectories that mix base- and teacher-preferred tokens. As training continues, more probability mass shifts toward teacher-preferred tokens, and the sampled trajectories increasingly follow those preferences.

Calibration controls how far this progression proceeds. Larger $\lambda$ more strongly preserves base-preferred probabilities, limiting the sampling of teacher-preferred alternatives and keeping the target closer to its base. Smaller $\lambda$ permits stronger redistribution and more extensive trajectory migration. This explains the higher target-policy--base overlap and lower virtual-teacher--target-policy overlap observed with larger coefficients.

\begin{table*}[ht]
\centering
\renewcommand{\arraystretch}{1.25}
\setlength{\tabcolsep}{3pt}

\providecommand{\gain}[1]{\scriptsize{\textcolor{gray}{(+#1)}}}
\providecommand{\bestgain}[1]{\scriptsize{\textcolor{teal}{\textbf{(+#1)}}}}

\vspace{-1.1em}
\caption{\textbf{Transferability of proxy update signals}.
Mean@16 results of math benchmarks.
\methodAbb{} (Proxy Expert 4B $\rightarrow$ Target Policy 1.7B $\rightarrow$ Target Policy 8B)
uses proxy base model Qwen3-4B to explore update signals, which are then
transferred sequentially to Qwen3-1.7B and Qwen3-8B.}
\label{tab:signal_transferability}

\resizebox{0.88\linewidth}{!}{%
\begin{tabular}{@{}l ccccc@{}}
\toprule
\textbf{Model}
& \textbf{AIME24}
& \textbf{AIME25}
& \makecell[c]{\textbf{HMMT25} \textbf{(Feb)}}
& \makecell[c]{\textbf{HMMT25} \textbf{(Nov)}}
& \textbf{Avg.} \\








\midrule
Qwen3-8B (Target Base)
& 24.4
& 22.5
& 12.3
& 10.0
& 17.3 \\



\rowcolor{tabcolor!20}
\methodAbb{} (Proxy Expert 1.7B $\rightarrow$ Target Policy 8B)
& 50.8 \gain{26.4}
& 42.1 \gain{19.6}
& 22.9 \gain{10.6}
& 30.6 \gain{20.6}
& 36.6 \gain{19.3} \\

\rowcolor{tabcolor!20}
\methodAbb{} (Proxy Expert 4B $\rightarrow$ Target Policy 8B)
& \textbf{62.5} \bestgain{38.1}
& \textbf{52.7} \bestgain{30.2}
& \textbf{34.2} \bestgain{21.9}
& \textbf{40.4} \bestgain{30.4}
& \textbf{47.5} \bestgain{30.2} \\

\rowcolor{tabcolor!20}
\methodAbb{} (Proxy Expert 4B $\rightarrow$ Target Policy 1.7B $\rightarrow$ Target Policy 8B)
& 60.6 \gain{36.2}
& 49.6 \gain{27.1}
& 28.5 \gain{16.2}
& 36.7 \gain{26.7}
& 43.9 \gain{26.6} \\

\bottomrule
\end{tabular}%
}

\vspace{-0.5em}
\end{table*}

\begin{table*}
\centering
\renewcommand{\arraystretch}{1.30}
\setlength{\tabcolsep}{3pt}
\providecommand{\gain}[1]{%
\scriptsize\textcolor{gray}{(+#1)}%
}
\providecommand{\bestgain}[1]{%
\scriptsize\textcolor{teal}{\textbf{(+#1)}}%
}
\caption{\textbf{Cross-generation reusability of transferrable update signals}.
Mean@16 results of math benchmarks. ``SW7B'' and ``Q8B'' denote
Skywork-OR1-Math-7B (proxy expert) and Qwen3-8B
(target policy), respectively. Update signals explored by an
older-generation proxy can effectively guide the training of a
new-generation target model.}
\label{tab:cross_gen_reusability}
\resizebox{0.80\linewidth}{!}{%
\begin{tabular}{@{}lccccc@{}}
\toprule
\makecell[l]{\textbf{Model}}
& \makecell[c]{\textbf{AIME24}}
& \makecell[c]{\textbf{AIME25}}
& \makecell[c]{\textbf{HMMT25 (Feb)}}
& \makecell[c]{\textbf{HMMT25 (Nov)}}
& \makecell[c]{\textbf{AVG.}} \\
\midrule
\multicolumn{6}{c}{%
\cellcolor{gray!15}\textit{Cross-generation Base Models}} \\
Qwen2.5-7B (Proxy Base)
& 4.4
& 0.8
& 0
& 1.5
& 1.7 \\
Qwen3-8B (Target Base)
& 24.4
& 22.5
& 12.3
& 10.0
& 17.3 \\
\midrule
\multicolumn{6}{c}{%
\cellcolor{gray!15}\textit{Obtaining Update Signals}} \\
Skywork-OR1-Math-7B
& 63.5 \bestgain{59.1}
& 48.5 \bestgain{47.7}
& 30.8 \bestgain{30.8}
& 37.7 \gain{36.2}
& 45.2 \bestgain{43.5} \\
Qwen3-8B Math (RL)
& 65.2 \gain{40.8}
& 53.8 \gain{31.3}
& 29.8 \gain{17.5}
& 29.4 \gain{19.4}
& 44.5 \gain{27.2} \\
\midrule
\multicolumn{6}{c}{%
\cellcolor{gray!15}%
\textit{Update Signal Transferred Across Generations}} \\
OPD (Proxy Expert SW7B $\rightarrow$ Target Policy Q8B)
& 60.4 \gain{36.0}
& 48.8 \gain{26.3}
& 29.0 \gain{16.7}
& 38.1 \gain{28.1}
& 44.1 \gain{26.8} \\
\rowcolor{tabcolor!20}
\methodAbb{} (Proxy Expert SW7B $\rightarrow$ Target Policy Q8B)
& \textbf{69.2} \gain{44.8}
& \textbf{57.5} \gain{35.0}
& \textbf{35.0} \gain{22.7}
& \textbf{48.1} \bestgain{38.1}
& \textbf{52.5} \gain{35.2} \\
\bottomrule
\end{tabular}%
}
\end{table*}

The calibration term therefore acts as a soft trust-region constraint around the target base. Together with Figure~\ref{fig:sensitivity_analysis}, these results show that effective transfer requires an appropriate degree of migration, not simply the closest match to the virtual teacher. The highest task scores occur at intermediate coefficients, and the preferred balance differs between proxies. This reflects a broader consideration for distribution-matching methods: their objectives guide policy migration rather than directly maximize task reward. Task-level feedback can therefore help identify the most useful degree of transfer through coefficient selection or adaptive calibration.

\subsection{Transferability}
\label{sec:transferability}

\subsubsection{Cross-Scale and Sequential Transfer}

We further evaluate proxy update signal transfer across different model scales to investigate its transferability and characterize its behavior. As shown in Table~\ref{tab:signal_transferability}, update signals explored by both the 1.7B and 4B proxy models transfer effectively to the 8B target model. In both cases, \methodAbb{} substantially improves the target model.

We further investigate whether update signals remain effective after multiple rounds of transfer. Specifically, we first transfer the update signals explored by Qwen3-4B to Qwen3-1.7B and then transfer them from Qwen3-1.7B to Qwen3-8B, using $\lambda=1.0$ at both stages. Despite this sequential transfer, Qwen3-8B still achieves substantial performance gains. Moreover, the resulting improvement approaches that of the direct 4B$\rightarrow$8B transfer, indicating that the explored update signals are largely preserved across multiple rounds of transfer.

\subsubsection{Cross-Generation Reusability}

Transferring update signals provides a natural way to reuse signals explored by earlier model generations to improve newer models. To evaluate this cross-generation reusability, we conduct transfer experiments between Qwen2.5-7B~\cite{qwen2025qwen25technicalreport} and Qwen3-8B. Skywork-OR1-Math-7B~\cite{he2025skywork} is an expert model derived from Qwen2.5-7B through post-training on multiple datasets with multiple training methods. We therefore treat Qwen2.5-7B and Skywork-OR1-Math-7B as the proxy base and proxy expert, respectively. For comparison, we also obtain a reference expert, Qwen3-8B (RL), by training Qwen3-8B (Target Base) on DeepMath-103K with GRPO for 400 steps.

As shown in Table~\ref{tab:cross_gen_reusability}, update signals explored by the older-generation proxy effectively improve the newer-generation target policy. Moreover, transferring these signals to Qwen3-8B yields a greater improvement than directly exploring update signals with the Qwen3-8B reference expert. In contrast, OPD, which directly aligns the target policy with the absolute output distribution of the older-generation proxy expert, performs substantially worse than \methodAbb{}. These results demonstrate that relative update signals remain reusable across model generations and can provide valuable guidance for improving newer target policies.

\subsection{Composability}
\label{sec:exp_multi_domain}

\subsubsection{Multi-Domain Joint Transfer}

To evaluate whether independently explored update signals can be combined, we jointly distill three Qwen3-1.7B proxy experts specializing in Math, Code, and instruction following (IF) into a single Qwen3-8B target policy. Following the multi-domain formulation in Section~\ref{subsec:multi_domain}, each training batch contains examples from only one domain and uses the corresponding proxy expert, with training alternating across the three domains. We evaluate both OPD and \methodAbb{} under this joint training setting and compare each method with its single-domain counterparts.

\begin{table*}[t]
\centering
\renewcommand{\arraystretch}{1.15}
\caption{\textbf{Single-domain and multi-domain distillation from 1.7B proxy experts to an 8B target policy.}}
\label{tab:multi_domain}

\resizebox{0.92\textwidth}{!}{%
\begin{tabular}{l ccccc cccc c}
\toprule
\multirow{2}{*}{\raisebox{-1ex}{\textbf{Model}}}
& \multicolumn{5}{c}{\textbf{Math}}
& \multicolumn{4}{c}{\textbf{Code}}
& \multicolumn{1}{c}{\textbf{IF}} \\
\cmidrule(lr){2-6}
\cmidrule(lr){7-10}
\cmidrule(lr){11-11}
& \textbf{AIME24}
& \textbf{AIME25}
& \makecell[c]{\textbf{HMMT25}\\[-1pt]\textbf{(FEB)}}
& \makecell[c]{\textbf{HMMT25}\\[-1pt]\textbf{(NOV)}}
& \textbf{AVG.}
& \textbf{TACO}
& \makecell[c]{\textbf{Code}\\[-1pt]\textbf{Forces}}
& \textbf{APPs}
& \textbf{AVG.}
& \textbf{Nemotron} \\
\midrule
\multicolumn{11}{c}{\cellcolor{gray!15}\textit{Single-Domain Distillation}} \\
OPD (Proxy Expert 1.7B $\rightarrow$ Target Policy 8B)
& 41.0 & 34.4 & 19.2 & 18.5 & 28.3
& 23.0 & 36.8 & 47.5 & 35.8 & \textbf{47.7} \\
\rowcolor{tabcolor!25}
\methodAbb{} (Proxy Expert 1.7B $\rightarrow$ Target Policy 8B)
& 50.8 & 42.1 & 22.9 & 30.6 & 36.6
& 26.0 & 40.2 & \textbf{51.2} & 39.2 & 45.1 \\
\midrule
\multicolumn{11}{c}{\cellcolor{gray!15}\textit{Multi-Domain Distillation}} \\
OPD (Proxy Expert 1.7B $\rightarrow$ Target Policy 8B)
& 39.6 & 35.8 & 20.4 & 18.8 & 28.7
& 21.2 & 35.6 & 42.1 & 33.0 & 36.2 \\
\rowcolor{tabcolor!25}
\methodAbb{} (Proxy Expert 1.7B $\rightarrow$ Target Policy 8B)
& \textbf{52.1} & \textbf{44.2} & \textbf{23.8} & \textbf{31.3} & \textbf{37.8}
& \textbf{27.4} & \textbf{45.3} & 50.4 & \textbf{41.0} & 38.8 \\
\bottomrule
\end{tabular}%
}
\vspace{-1em}
\end{table*}

\begin{figure}[t]
\centering
\begin{subfigure}[t]{0.44\linewidth}
    \centering
    \includegraphics[width=\linewidth]
        {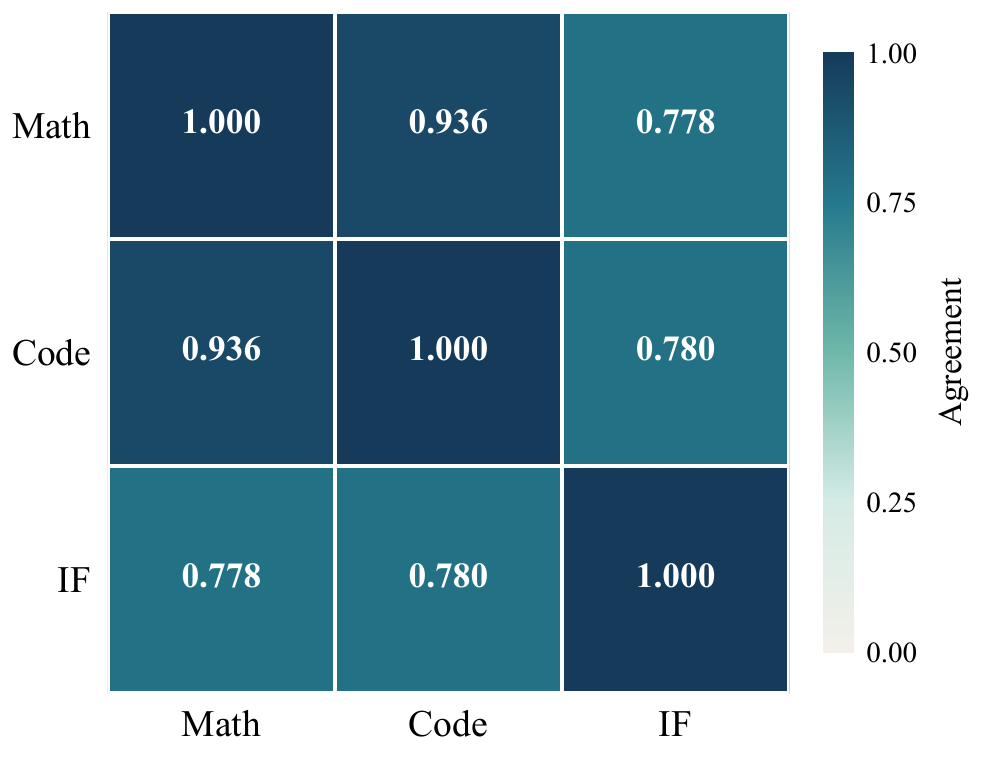}
    \caption{Sampled token agreement}
    \label{fig:teacher_agreement_top1}
\end{subfigure}%
\hspace{0.04\linewidth}%
\begin{subfigure}[t]{0.44\linewidth}
    \centering
    \includegraphics[width=\linewidth]
        {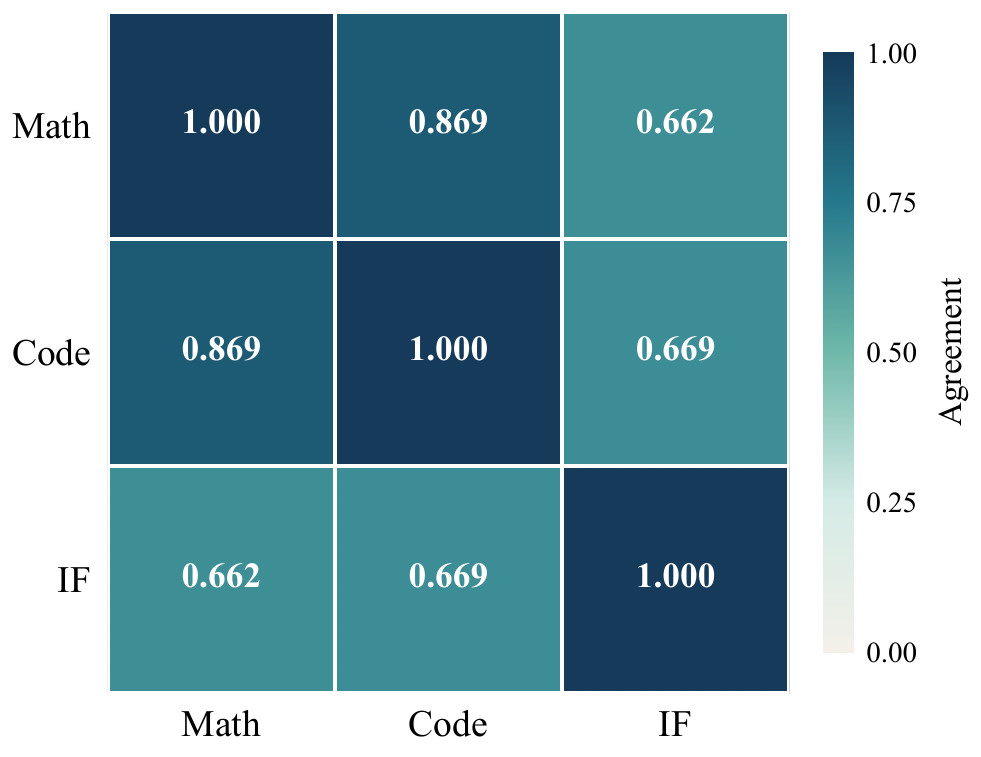}
    \caption{Top-64 tokens agreement}
    \label{fig:teacher_agreement_top64}
\end{subfigure}
\caption{\textbf{Token agreement between domain proxy experts.}
Sampled-token and top-64 agreement are measured via teacher forcing on shared prefixes of proxy-generated trajectories and averaged across sampled trajectories.}
\label{fig:teacher_agreement}
\vspace{-1em}
\end{figure}

As shown in Table~\ref{tab:multi_domain}, joint \methodAbb{} achieves higher domain-average scores than joint OPD in Math (37.8 vs.\ 28.7), Code (41.0 vs.\ 33.0), and IF (38.8 vs.\ 36.2). Relative to single-domain \methodAbb{}, the Math and Code averages increase from 36.6 to 37.8 and from 39.2 to 41.0, respectively, whereas the IF score decreases from 45.1 to 38.8. These results demonstrate that multiple domain-specific signals can jointly guide one target model, but the benefits are not uniform across domains. The Math and Code gains are consistent with positive transfer, while the IF decline indicates a trade-off under joint training.

\subsubsection{Signal Compatibility and Domain Interference}

To explore a possible explanation for these domain-dependent outcomes, we examine agreement between the domain proxy experts. We evaluate each expert by teacher forcing on trajectories generated by the other proxy experts and compare their token predictions under shared prefixes. This diagnostic measures agreement between the experts' output distributions.

Figure~\ref{fig:teacher_agreement} shows that the Math and Code proxy experts have high token agreement (0.936 for sampled tokens and 0.869 for top-64 tokens), whereas the IF proxy expert agrees less with either Math (0.778 and 0.662, respectively) or Code (0.780 and 0.669, respectively). This agreement pattern is consistent with the Math and Code gains and the IF decline under joint transfer.

These observations provide explanatory evidence for a possible role of guidance compatibility. Taken together, these experiments support composability in the sense that independently explored domain-specific signals can be jointly transferred to a shared target policy. Composability does not imply uniformly beneficial composition: the tested combination improves Math and Code relative to single-domain transfer but sacrifices some IF performance, motivating closer examination of signal compatibility when combining domain-specific supervision.

\subsection{Efficiency}
\label{exp:efficiency}

\subsubsection{Exploration Budget vs.\ Signal Quality}


\begin{figure}[t]
\centering
\begin{subfigure}[t]{0.48\linewidth}
    \centering
    \includegraphics[width=\linewidth]{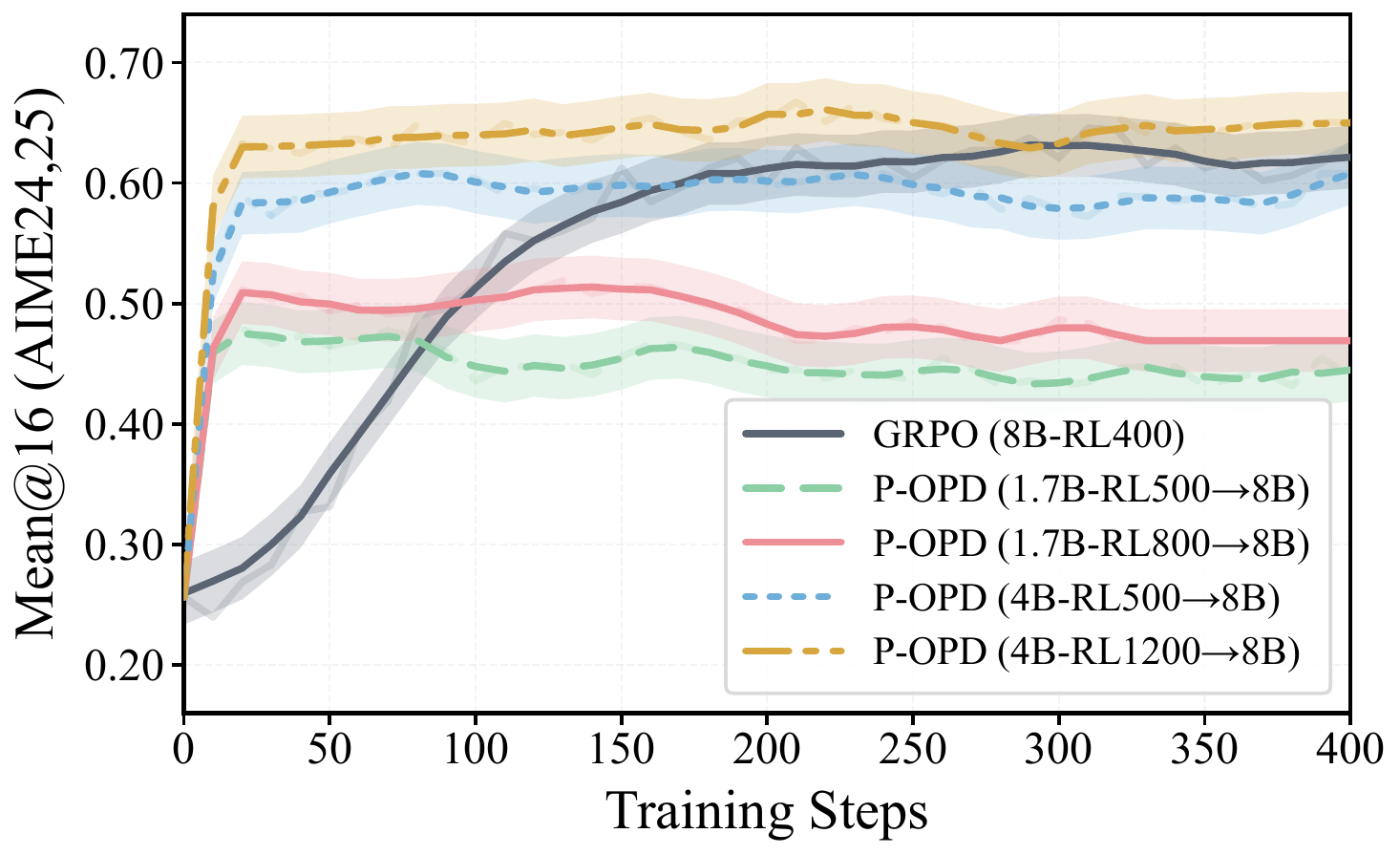}
    \caption{Mean@16 trajectories}
    \label{fig:mean16_trajectories}
\end{subfigure}
\hfill
\begin{subfigure}[t]{0.48\linewidth}
    \centering
    \includegraphics[width=\linewidth]{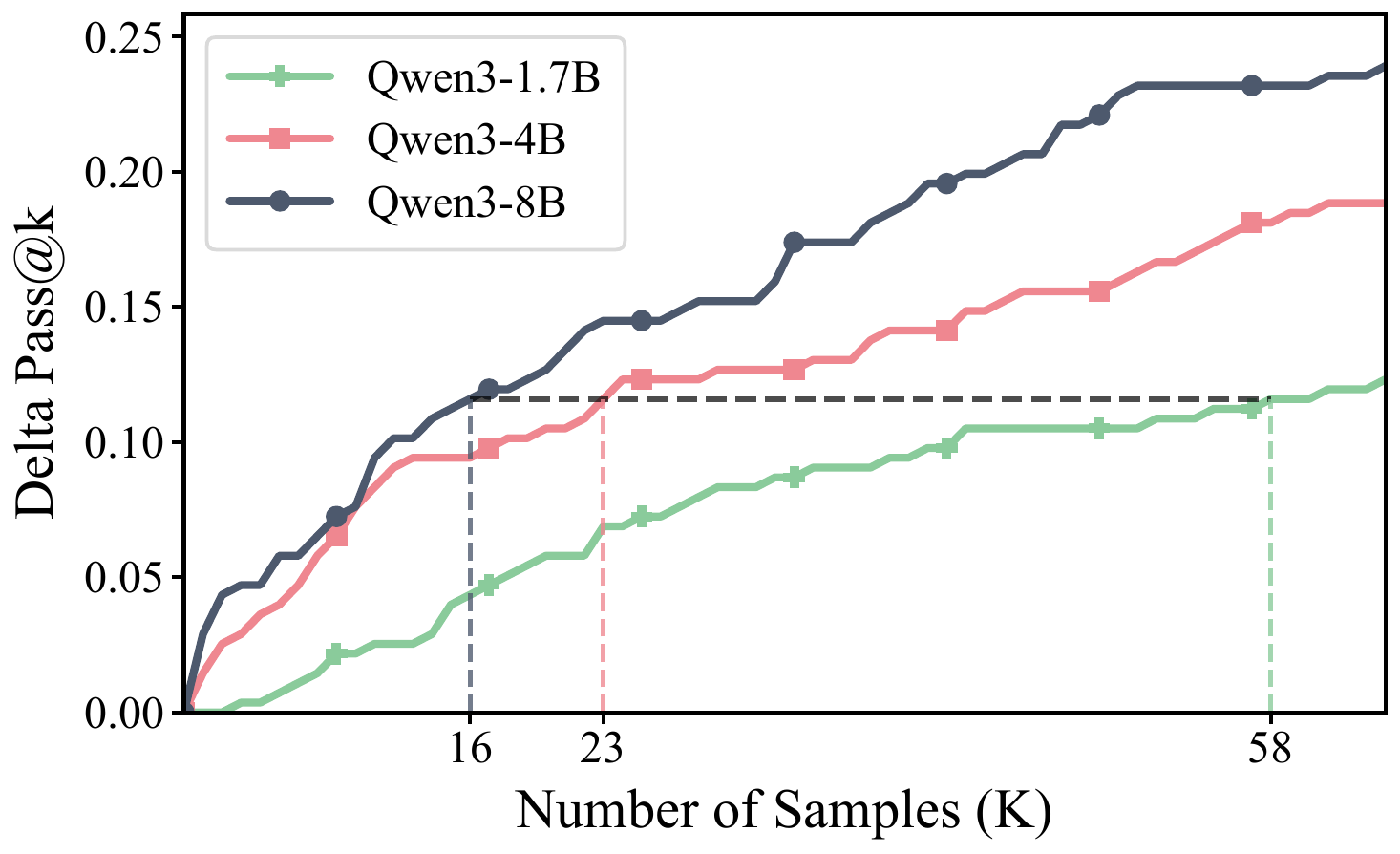}
    \caption{$\Delta$ Pass@K versus the sampling budget K.}
    \label{fig:delta_passK}
\end{subfigure}
\caption{All \methodAbb{} runs converge within 70 steps and are extended using converged-region statistics. More samples consistently uncover additional correct solutions.}
\label{fig:training-and-sampling}
\end{figure}

Proxy-guided transfer offloads update-signal exploration to smaller models, but the explored signals may differ in quality from those of the target policy. To quantify this difference, we use Qwen3-8B as the target model and Qwen3-1.7B and Qwen3-4B as proxies with varying exploration budgets. In Figure~\ref{fig:mean16_trajectories}, GRPO (8B-RL400) denotes direct exploration from the Qwen3-8B target base to its reference expert for 400 steps, while \methodAbb{} (1.7B-RL500$\rightarrow$8B) denotes 500-step proxy exploration by Qwen3-1.7B followed by signal transfer to the Qwen3-8B target policy. Other settings follow the same notation.

Signal transfer converges in substantially fewer target-model updates than direct exploration. Under comparable exploration step counts (400 steps for Qwen3-8B versus 500 steps for the proxy models), Qwen3-8B explores higher-quality signals than the smaller proxies. However, increasing the proxy exploration budget narrows this gap. As shown in Figure~\ref{fig:mean16_trajectories}, \methodAbb{} (4B-RL1200$\rightarrow$8B) converges slightly above GRPO (8B-RL400). This result indicates that more extensive proxy exploration can improve the quality of the transferred signal.

Figure~\ref{fig:delta_passK} examines the additional correct solutions uncovered as the sampling budget increases. The gains for smaller models motivate allocating more samples to proxy exploration. Together, these observations suggest two ways to invest in signal quality: extending proxy training and increasing sampling during exploration.

\subsubsection{Training Cost and Reuse Amortization}


\begin{wrapfigure}{r}{0.54\textwidth}
\vspace{-1.2em}
\centering
\includegraphics[width=0.88\linewidth]{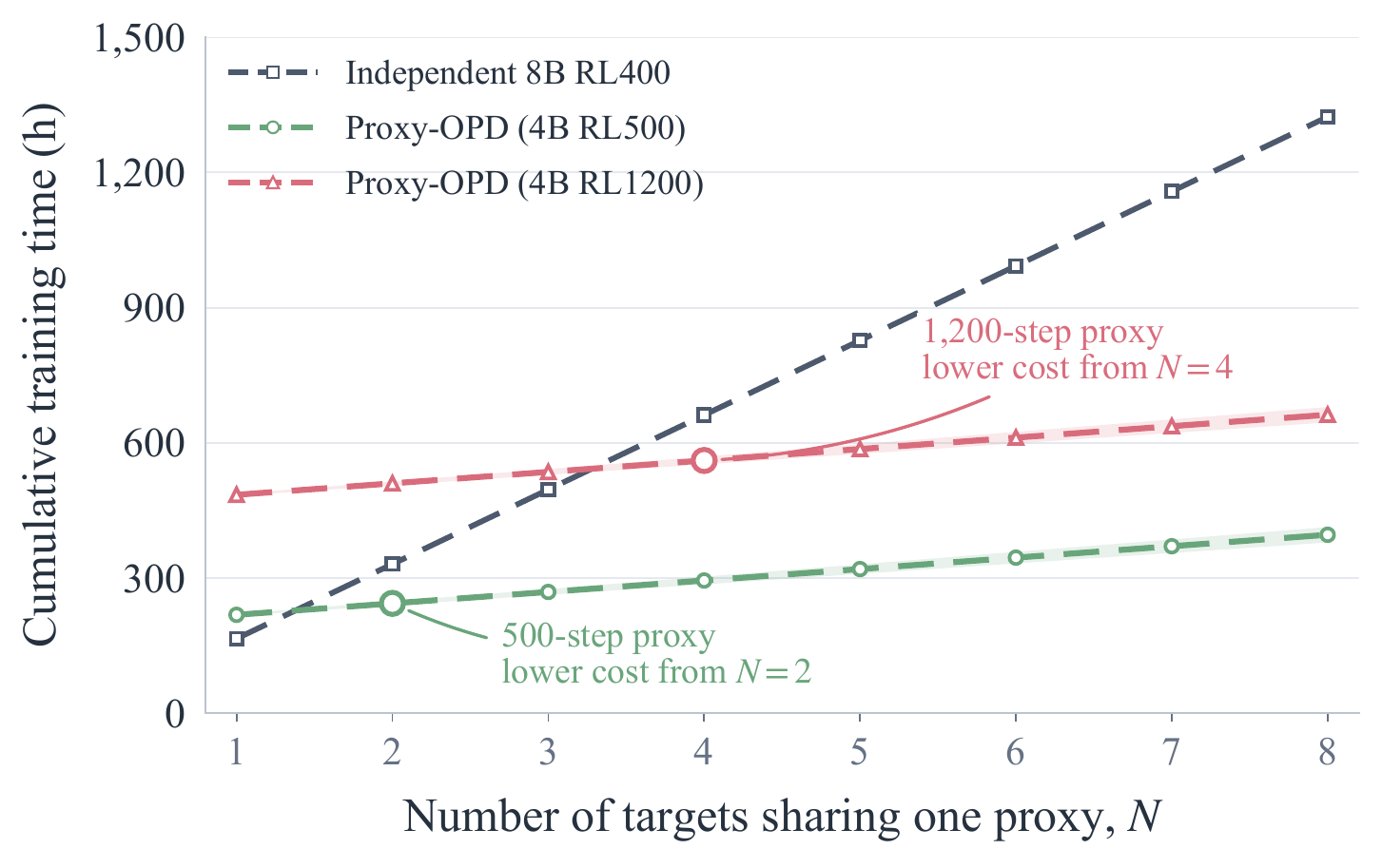}
\caption{\textbf{Exploration cost amortization through signal reuse.}
Projected cumulative training time for $N$ compatible 8B targets. Independent RL repeats 400-step exploration for each target, whereas \methodAbb{} shares one proxy exploration run and adds only signal-transfer time for each target.}
\label{fig:exploration_cost_amortization}
\end{wrapfigure}

We compare per-step exploration costs on the same eight A100 GPUs, using identical training hyperparameters, including batch and rollout group sizes. The 4B proxy takes 23.2 min per step, compared with 24.8 min for the 8B target, indicating a modest difference in exploration speed.

The main efficiency gain comes from reusing exploration outcomes. Once a proxy has explored the training data, its frozen base--expert pair provides relative update signals on each target's own trajectories. New targets can absorb these signals without repeating reward-guided exploration on the same data. Across repeated runs, 4B-to-8B signal transfer takes $25.37\pm2.14$ h (mean $\pm$ standard deviation), approximately 15\% of the 165.43 h required for direct 8B-RL400 training.

For $N$ compatible targets, each with the measured 8B training workload, the projected cumulative costs can be described as:
\begin{equation}
    \begin{aligned}
        C_{\mathrm{direct}}(N) &= N T_{\mathrm{RL}}, \\
        C_{\mathrm{proxy}}(N) &= T_{\mathrm{explore}} + N T_{\mathrm{distill}}.
    \end{aligned}
    \label{eq:exploration_amortization}
\end{equation}
where $T_{\mathrm{RL}}$ is the direct target exploration time, $T_{\mathrm{explore}}$ is the one-time proxy exploration cost, and $T_{\mathrm{distill}}$ is the per-target signal-transfer time. The measured 4B exploration costs are 192.99 h for 500 steps and 459.34 h for 1,200 steps. As shown in Figure~\ref{fig:exploration_cost_amortization}, sharing these runs yields lower cumulative cost than independent 8B-RL400 training from $N=2$ and $N=4$, respectively. For four targets, the 1,200-step proxy pipeline requires approximately 560.8 h, compared with 661.7 h for independent target exploration.

Reuse also makes larger exploration budgets economical: the exploration cost per target decreases as $T_{\mathrm{explore}}/N$. This connects directly to Figure~\ref{fig:mean16_trajectories}, where extending 4B exploration to 1,200 steps improves transfer quality and yields performance slightly above direct 8B-RL400. Proxy selection can thus prioritize exploration quality rather than model size: smaller, similarly sized, or larger models can all provide reusable update signals. Investing in longer training or broader sampling with a suitable proxy makes fuller use of the same data, while subsequent targets absorb the resulting improvements through low-cost signal transfer.

\begin{figure*}[t]
    \centering
    \includegraphics[width=0.88\linewidth]{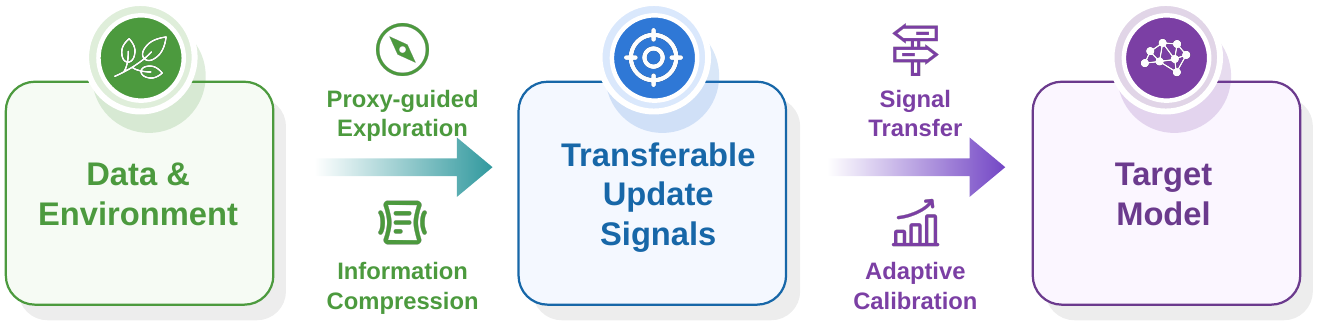}
    \caption{\textbf{A forward-looking, signal-centric view of post-training.}
    Proxy exploration consolidates useful information from data and environments into relative update signals represented by a proxy base--expert pair. These signals guide target-model training, with adaptive calibration as a key research direction for improving their absorption.}
    \label{fig:discussion}
\end{figure*}

\section{Discussion and Future Work}

\methodAbb{} motivates a signal-centric view of future post-training, organized around discovering useful update signals and enabling target models to absorb them (\cref{fig:discussion}). From this perspective, proxy exploration acts as \emph{information compression}: reward-guided training consolidates useful information from post-training data and environments into relative policy updates represented by a reusable proxy base--expert pair. Target models can then absorb these improvements without repeating reward-guided exploration on the same data. Our exploration-budget and amortization results motivate selecting proxies for exploration quality rather than model size and investing in more extensive exploration whose cost can be shared across targets. More effective exploration strategies and sampling-budget allocation are promising directions for extracting higher-quality supervision from the available data.

Effective signal absorption is the complementary research direction. Our sensitivity and distribution-migration analyses show how calibration balances signal absorption against departure from the target base. Building on our fixed-coefficient formulation, \emph{adaptive calibration} could adjust transfer strength along target-generated trajectories according to local differences between proxy and target policies. The domain-dependent gains and instruction-following decline in multi-domain transfer also motivate interference-aware aggregation of multiple proxy signals. Broader evaluation beyond the Qwen model family would further characterize signal transferability. Together, these directions aim to improve both the quality of explored signals and their absorption by different target models.

\section{Conclusion}

We introduced \methodFull{} (\methodAbb{}), an OPD-style post-training strategy that transfers reward-induced relative policy updates rather than matching an expert's absolute output distribution. By extracting changes between proxy base and expert policies and retaining the target base as a reference, \methodAbb{} decouples proxy exploration from target-model training. Experiments across mathematical reasoning, code generation, and instruction following show improvements over target base models and conventional OPD in most evaluated settings. The signals remain useful across model scales, generations, and sequential transfer, while multi-domain transfer combines expert guidance with domain-dependent trade-offs. In some Math settings, increased proxy exploration yields transferred performance comparable to direct target-model RL. Measured phase costs and conditional amortization analysis show how compatible-target reuse can share that exploration investment, without implying lower end-to-end cost for every single target. Together, these findings support relative policy updates as reusable and calibratable supervision for an ``explore once, transfer multiple times'' post-training paradigm.



{
\bibliographystyle{unsrt}  
\bibliography{preprint}

@article{yang2026learning,
  title={Learning beyond Teacher: Generalized On-Policy Distillation with Reward Extrapolation},
  author={Yang, Wenkai and Liu, Weijie and Xie, Ruobing and Yang, Kai and Yang, Saiyong and Lin, Yankai},
  journal={arXiv preprint arXiv:2602.12125},
  year={2026},
  url={https://arxiv.org/abs/2602.12125}
}

@article{schulman2017proximal,
  title={Proximal Policy Optimization Algorithms},
  author={Schulman, John and Wolski, Filip and Dhariwal, Prafulla and Radford, Alec and Klimov, Oleg},
  journal={arXiv preprint arXiv:1707.06347},
  year={2017},
  doi={10.48550/arXiv.1707.06347},
  url={https://arxiv.org/abs/1707.06347}
}

@article{shao2024deepseekmath,
  title={{DeepSeekMath}: Pushing the Limits of Mathematical Reasoning in Open Language Models},
  author={Shao, Zhihong and Wang, Peiyi and Zhu, Qihao and Xu, Runxin and Song, Junxiao and Bi, Xiao and Zhang, Haowei and Zhang, Mingchuan and Li, Y. K. and Wu, Y. and Guo, Daya},
  journal={arXiv preprint arXiv:2402.03300},
  year={2024},
  doi={10.48550/arXiv.2402.03300},
  url={https://arxiv.org/abs/2402.03300}
}

@inproceedings{agarwal2024policy,
  title={On-Policy Distillation of Language Models: Learning from Self-Generated Mistakes},
  author={Agarwal, Rishabh and Vieillard, Nino and Zhou, Yongchao and Stanczyk, Piotr and Ramos Garea, Sabela and Geist, Matthieu and Bachem, Olivier},
  booktitle={The Twelfth International Conference on Learning Representations},
  year={2024},
  url={https://openreview.net/forum?id=3zKtaqxLhW}
}

@inproceedings{lai2025survey,
  title={A survey of post-training scaling in large language models},
  author={Lai, Hanyu and Liu, Xiao and Gao, Junjie and Cheng, Jiale and Qi, Zehan and Xu, Yifan and Yao, Shuntian and Zhang, Dan and Du, Jinhua and Hou, Zhenyu and others},
  booktitle={Proceedings of the 63rd Annual Meeting of the Association for Computational Linguistics (Volume 1: Long Papers)},
  pages={2771--2791},
  year={2025}
}

@inproceedings{lightman2024let,
  title={Let's Verify Step by Step},
  author={Lightman, Hunter and Kosaraju, Vineet and Burda, Yuri and Edwards, Harrison and Baker, Bowen and Lee, Teddy and Leike, Jan and Schulman, John and Sutskever, Ilya and Cobbe, Karl},
  booktitle={The Twelfth International Conference on Learning Representations},
  year={2024},
  url={https://openreview.net/forum?id=v8L0pN6EOi}
}

@article{lee2023rlaif,
  title={Rlaif: Scaling reinforcement learning from human feedback with ai feedback},
  author={Lee, Harrison and Phatale, Samrat and Mansoor, Hassan and Lu, Kellie Ren and Mesnard, Thomas and Ferret, Johan and Bishop, Colton and Hall, Ethan and Carbune, Victor and Rastogi, Abhinav},
  year={2023}
}

@inproceedings{liu2024tuning,
  title={Tuning Language Models by Proxy},
  author={Liu, Alisa and Han, Xiaochuang and Wang, Yizhong and Tsvetkov, Yulia and Choi, Yejin and Smith, Noah A.},
  booktitle={First Conference on Language Modeling},
  year={2024},
  url={https://openreview.net/forum?id=dribhnhm1i}
}

@article{feng2026weak,
  title={Weak-to-Strong Generalization via Direct On-Policy Distillation},
  author={Feng, Shiyuan and Gao, Huan-ang and Chi, Haohan and Wu, Hanlin and Zhang, Zhilong and Jiang, Zheng and He, Bingxiang and Ma, Wei-Ying and Zhang, Ya-Qin and Zhou, Hao},
  journal={arXiv preprint arXiv:2607.05394},
  year={2026},
  url={https://arxiv.org/abs/2607.05394}
}

@inproceedings{rafailov2023direct,
  title={Direct Preference Optimization: Your Language Model is Secretly a Reward Model},
  author={Rafailov, Rafael and Sharma, Archit and Mitchell, Eric and Manning, Christopher D. and Ermon, Stefano and Finn, Chelsea},
  booktitle={Advances in Neural Information Processing Systems},
  volume={36},
  pages={53728--53741},
  year={2023},
  doi={10.52202/075280-2338},
  url={https://proceedings.neurips.cc/paper_files/paper/2023/hash/a85b405ed65c6477a4fe8302b5e06ce7-Abstract-Conference.html}
}

@article{yang2025qwen3,
  title={Qwen3 technical report},
  author={Yang, An and Li, Anfeng and Yang, Baosong and Zhang, Beichen and Hui, Binyuan and Zheng, Bo and Yu, Bowen and Gao, Chang and Huang, Chengen and Lv, Chenxu and others},
  journal={arXiv preprint arXiv:2505.09388},
  year={2025}
}

@article{balunovic2026matharena,
  title={Matharena: Evaluating llms on uncontaminated math competitions},
  author={Balunovic, Mislav and Dekoninck, Jasper and Petrov, Ivo and Jovanovi{\'c}, Nikola and Vechev, Martin},
  journal={Advances in Neural Information Processing Systems},
  volume={38},
  year={2026}
}

@article{he2025deepmath,
  title={Deepmath-103k: A large-scale, challenging, decontaminated, and verifiable mathematical dataset for advancing reasoning},
  author={He, Zhiwei and Liang, Tian and Xu, Jiahao and Liu, Qiuzhi and Chen, Xingyu and Wang, Yue and Song, Linfeng and Yu, Dian and Liang, Zhenwen and Wang, Wenxuan and others},
  journal={arXiv preprint arXiv:2504.11456},
  year={2025}
}

@article{cui2025process,
  title={Process reinforcement through implicit rewards},
  author={Cui, Ganqu and Yuan, Lifan and Wang, Zefan and Wang, Hanbin and Zhang, Yuchen and Chen, Jiacheng and Li, Wendi and He, Bingxiang and Fan, Yuchen and Yu, Tianyu and others},
  journal={arXiv preprint arXiv:2502.01456},
  year={2025}
}

@article{li2024numinamath,
  title={Numinamath: The largest public dataset in ai4maths with 860k pairs of competition math problems and solutions},
  author={Li, Jia and Beeching, Edward and Tunstall, Lewis and Lipkin, Ben and Soletskyi, Roman and Huang, Shengyi and Rasul, Kashif and Yu, Longhui and Jiang, Albert Q and Shen, Ziju and others},
  journal={Hugging Face repository},
  volume={13},
  number={9},
  pages={9},
  year={2024}
}

@misc{2023opencompass,
    title={OpenCompass: A Universal Evaluation Platform for Foundation Models},
    author={OpenCompass Contributors},
    howpublished = {\url{https://github.com/open-compass/opencompass}},
    year={2023}
}

@article{sheng2024hybridflow,
  title   = {HybridFlow: A Flexible and Efficient RLHF Framework},
  author  = {Guangming Sheng and Chi Zhang and Zilingfeng Ye and Xibin Wu and Wang Zhang and Ru Zhang and Yanghua Peng and Haibin Lin and Chuan Wu},
  year    = {2024},
  journal = {arXiv preprint arXiv: 2409.19256}
}

@inproceedings{yoon2024tlcr,
  title={{TLCR}: Token-Level Continuous Reward for Fine-grained Reinforcement Learning from Human Feedback},
  author={Yoon, Eunseop and Yoon, Hee Suk and Eom, SooHwan and Han, Gunsoo and Nam, Daniel and Jo, Daejin and On, Kyoung-Woon and Hasegawa-Johnson, Mark and Kim, Sungwoong and Yoo, Chang},
  booktitle={Findings of the Association for Computational Linguistics: ACL 2024},
  publisher={Association for Computational Linguistics},
  pages={14969--14981},
  year={2024},
  doi={10.18653/v1/2024.findings-acl.889},
  url={https://aclanthology.org/2024.findings-acl.889/}
}

@inproceedings{kim2016sequence,
  title={Sequence-Level Knowledge Distillation},
  author={Kim, Yoon and Rush, Alexander M.},
  booktitle={Proceedings of the 2016 Conference on Empirical Methods in Natural Language Processing},
  publisher={Association for Computational Linguistics},
  pages={1317--1327},
  year={2016},
  doi={10.18653/v1/D16-1139},
  url={https://aclanthology.org/D16-1139/}
}

@article{busbridge2025distillation,
  title={Distillation scaling laws},
  author={Busbridge, Dan and Shidani, Amitis and Weers, Floris and Ramapuram, Jason and Littwin, Etai and Webb, Russ},
  journal={arXiv preprint arXiv:2502.08606},
  year={2025}
}

@article{song2026survey,
  title={A survey of on-policy distillation for large language models},
  author={Song, Mingyang and Zheng, Mao},
  journal={arXiv preprint arXiv:2604.00626},
  year={2026}
}

@article{hinton2015distilling,
  title={Distilling the Knowledge in a Neural Network},
  author={Hinton, Geoffrey and Vinyals, Oriol and Dean, Jeff},
  journal={arXiv preprint arXiv:1503.02531},
  year={2015},
  doi={10.48550/arXiv.1503.02531},
  url={https://arxiv.org/abs/1503.02531}
}

@inproceedings{zhu2025weak,
  title={Weak-to-Strong Preference Optimization: Stealing Reward from Weak Aligned Model},
  author={Zhu, Wenhong and He, Zhiwei and Wang, Xiaofeng and Liu, Pengfei and Wang, Rui},
  booktitle={The Thirteenth International Conference on Learning Representations},
  year={2025},
  url={https://openreview.net/forum?id=f7KxfUrRSb}
}

@misc{deepseekai2026deepseekv4highlyefficientmilliontoken,
      title={DeepSeek-V4: Towards Highly Efficient Million-Token Context Intelligence}, 
      author={DeepSeek-AI},
      year={2026},
      eprint={2606.19348},
      archivePrefix={arXiv},
      primaryClass={cs.CL},
      url={https://arxiv.org/abs/2606.19348}, 
}

@misc{qwen2025qwen25technicalreport,
      title={Qwen2.5 Technical Report}, 
      author={Qwen and : and An Yang and Baosong Yang and Beichen Zhang and Binyuan Hui and Bo Zheng and Bowen Yu and Chengyuan Li and Dayiheng Liu and Fei Huang and Haoran Wei and Huan Lin and Jian Yang and Jianhong Tu and Jianwei Zhang and Jianxin Yang and Jiaxi Yang and Jingren Zhou and Junyang Lin and Kai Dang and Keming Lu and Keqin Bao and Kexin Yang and Le Yu and Mei Li and Mingfeng Xue and Pei Zhang and Qin Zhu and Rui Men and Runji Lin and Tianhao Li and Tianyi Tang and Tingyu Xia and Xingzhang Ren and Xuancheng Ren and Yang Fan and Yang Su and Yichang Zhang and Yu Wan and Yuqiong Liu and Zeyu Cui and Zhenru Zhang and Zihan Qiu},
      year={2025},
      eprint={2412.15115},
      archivePrefix={arXiv},
      primaryClass={cs.CL},
      url={https://arxiv.org/abs/2412.15115}, 
}

@article{he2025skywork,
  title={Skywork Open Reasoner 1 Technical Report},
  author={He, Jujie and Liu, Jiacai and Liu, Chris Yuhao and Yan, Rui and Wang, Chaojie and Cheng, Peng and Zhang, Xiaoyu and Zhang, Fuxiang and Xu, Jiacheng and Shen, Wei and Li, Siyuan and Zeng, Liang and Wei, Tianwen and Cheng, Cheng and An, Bo and Liu, Yang and Zhou, Yahui},
  journal={arXiv preprint arXiv:2505.22312},
  year={2025}
}

@misc{nvidia_nemotron_3_ultra_2026,
  title  = {Nemotron 3 Ultra: Open, Efficient Mixture-of-Experts Hybrid Mamba-Transformer Model for Agentic Reasoning},
  author = {{NVIDIA}},
  year   = {2026},
  url    = {https://research.nvidia.com/labs/nemotron/files/NVIDIA-Nemotron-3-Ultra-Technical-Report.pdf},
  note   = {White Paper}
}

@article{li2023taco,
  title={TACO: Topics in Algorithmic COde generation dataset},
  author={Rongao Li and Jie Fu and Bo-Wen Zhang and Tao Huang and Zhihong Sun and Chen Lyu and Guang Liu and Zhi Jin and Ge Li},
  journal={arXiv preprint arXiv:2312.14852},
  year={2023}
}

@article{hendrycksapps2021,
  title={Measuring Coding Challenge Competence With APPS},
  author={Dan Hendrycks and Steven Basart and Saurav Kadavath and Mantas Mazeika and Akul Arora and Ethan Guo and Collin Burns and Samir Puranik and Horace He and Dawn Song and Jacob Steinhardt},
  journal={NeurIPS},
  year={2021}
}

@misc{quan2025codeelobenchmarkingcompetitionlevelcode,
      title={CodeElo: Benchmarking Competition-level Code Generation of LLMs with Human-comparable Elo Ratings}, 
      author={Shanghaoran Quan and Jiaxi Yang and Bowen Yu and Bo Zheng and Dayiheng Liu and An Yang and Xuancheng Ren and Bofei Gao and Yibo Miao and Yunlong Feng and Zekun Wang and Jian Yang and Zeyu Cui and Yang Fan and Yichang Zhang and Binyuan Hui and Junyang Lin},
      year={2025},
      eprint={2501.01257},
      archivePrefix={arXiv},
      primaryClass={cs.CL},
      url={https://arxiv.org/abs/2501.01257}, 
}

@article{ouyang2022training,
  title={Training language models to follow instructions with human feedback},
  author={Ouyang, Long and Wu, Jeffrey and Jiang, Xu and Almeida, Diogo and Wainwright, Carroll and Mishkin, Pamela and Zhang, Chong and Agarwal, Sandhini and Slama, Katarina and Ray, Alex and others},
  journal={Advances in neural information processing systems},
  volume={35},
  pages={27730--27744},
  year={2022}
}

@misc{heo2026onpolicydeltadistillation,
      title={On-Policy Delta Distillation}, 
      author={Byeongho Heo and Jaehui Hwang and Sangdoo Yun and Dongyoon Han},
      year={2026},
      eprint={2607.15161},
      archivePrefix={arXiv},
      primaryClass={cs.LG},
      url={https://arxiv.org/abs/2607.15161}, 
}

@article{guo2025deepseek,
  title={Deepseek-r1: Incentivizing reasoning capability in llms via reinforcement learning},
  author={Guo, Daya and Yang, Dejian and Zhang, Haowei and Song, Junxiao and Wang, Peiyi and Zhu, Qihao and Xu, Runxin and Zhang, Ruoyu and Ma, Shirong and Bi, Xiao and others},
  journal={arXiv preprint arXiv:2501.12948},
  year={2025}
}

@inproceedings{gu2024minillm,
  title={Minillm: Knowledge distillation of large language models},
  author={Gu, Yuxian and Dong, Li and Wei, Furu and Huang, Minlie},
  booktitle={International Conference on Learning Representations},
  volume={2024},
  pages={32694--32717},
  year={2024}
}

@article{mitchell2023emulator,
  title={An emulator for fine-tuning large language models using small language models},
  author={Mitchell, Eric and Rafailov, Rafael and Sharma, Archit and Finn, Chelsea and Manning, Christopher D},
  journal={arXiv preprint arXiv:2310.12962},
  year={2023}
}

@article{christiano2017deep,
  title={Deep reinforcement learning from human preferences},
  author={Christiano, Paul F and Leike, Jan and Brown, Tom and Martic, Miljan and Legg, Shane and Amodei, Dario},
  journal={Advances in neural information processing systems},
  volume={30},
  year={2017}
}

@article{yu2026dapo,
  title={Dapo: An open-source llm reinforcement learning system at scale},
  author={Yu, Qiying and Zhang, Zheng and Zhu, Ruofei and Yuan, Yufeng and Zuo, Xiaochen and Yue, Yu and Dai, Weinan and Fan, Tiantian and Liu, Gaohong and Liu, Lingjun and others},
  journal={Advances in Neural Information Processing Systems},
  volume={38},
  pages={113222--113244},
  year={2026}
}

@inproceedings{azar2024general,
  title={A general theoretical paradigm to understand learning from human preferences},
  author={Azar, Mohammad Gheshlaghi and Guo, Zhaohan Daniel and Piot, Bilal and Munos, Remi and Rowland, Mark and Valko, Michal and Calandriello, Daniele},
  booktitle={International conference on artificial intelligence and statistics},
  pages={4447--4455},
  year={2024},
  organization={PMLR}
}

@inproceedings{ethayarajh2024model,
  title={Model alignment as prospect theoretic optimization},
  author={Ethayarajh, Kawin and Xu, Winnie and Muennighoff, Niklas and Jurafsky, Dan and Kiela, Douwe},
  booktitle={Forty-first International Conference on Machine Learning},
  year={2024}
}

@inproceedings{hong2024orpo,
  title={Orpo: Monolithic preference optimization without reference model},
  author={Hong, Jiwoo and Lee, Noah and Thorne, James},
  booktitle={Proceedings of the 2024 Conference on Empirical Methods in Natural Language Processing},
  pages={11170--11189},
  year={2024}
}

@inproceedings{wen2023f,
  title={F-divergence minimization for sequence-level knowledge distillation},
  author={Wen, Yuqiao and Li, Zichao and Du, Wenyu and Mou, Lili},
  booktitle={Proceedings of the 61st Annual Meeting of the Association for Computational Linguistics (Volume 1: Long Papers)},
  pages={10817--10834},
  year={2023}
}

@article{ko2024distillm,
  title={Distillm: Towards streamlined distillation for large language models},
  author={Ko, Jongwoo and Kim, Sungnyun and Chen, Tianyi and Yun, Se-Young},
  journal={arXiv preprint arXiv:2402.03898},
  year={2024}
}

@inproceedings{ross2011reduction,
  title={A reduction of imitation learning and structured prediction to no-regret online learning},
  author={Ross, St{\'e}phane and Gordon, Geoffrey and Bagnell, Drew},
  booktitle={Proceedings of the fourteenth international conference on artificial intelligence and statistics},
  pages={627--635},
  year={2011},
  organization={JMLR Workshop and Conference Proceedings}
}
}

\newgeometry{
  textheight=9in, textwidth=5.5in, top=1in,
  headheight=12pt, headsep=25pt, footskip=30pt
}

\appendix
\renewcommand{\thesection}{\Alph{section}}
\setcounter{section}{0}

\noindent{\LARGE\textbf{Appendix}\par}\normalsize

\vspace{10pt}
{\large \textbf{Contents}}
\startcontents[appendices]
\printcontents[appendices]{l}{1}{\setcounter{tocdepth}{3}}

\section{Experiment Details}
\subsection{Implementation Details}
\label{app:impl_details}

\paragraph{Models and framework.}
Our target policies and in-house proxy exploration runs use the Qwen3 family~\cite{yang2025qwen3} (1.7B, 4B, 8B) in non-thinking mode. The cross-generation experiment additionally uses Qwen2.5-7B and Skywork-OR1-Math-7B as an existing proxy base--expert pair, as described in Section~\ref{sec:effectiveness}. Our implementation is built upon verl~\cite{sheng2024hybridflow}, and all training runs performed in this study use 8$\times$NVIDIA A100 80GB.

\paragraph{Proxy exploration.}
The in-house proxy experts are obtained by training the corresponding proxy base policies with GRPO; Math and Code hyper-parameters are listed in Table~\ref{tab:grpo_hyperparams}. We train on DeepMath-103K~\cite{he2025deepmath} for the math domain (filtered following prior work) and on Eurus-RL-Code~\cite{cui2025process} for the code domain, using a rule-based verifiable reward (1 for correct answers and 0 otherwise).

\paragraph{Signal transfer (P-OPD).}
The training hyper-parameters for Math and Code signal transfer are summarized in Table~\ref{tab:pust_hyperparams_math}. The transfer settings in Sections~\ref{sec:effectiveness} specify the fixed calibration coefficients; sensitivity to $\lambda$ is analyzed in Section~\ref{sec:effectiveness} and Appendix~\ref{app:lambda_ablation}.

\paragraph{Evaluation protocol.}
For math benchmarks, we report Mean@16 (the average accuracy over 16 independently sampled responses) across a comprehensive suite of eight datasets: AIME24~\cite{li2024numinamath}, AIME25~\cite{2023opencompass}, AIME26, CMIMC25, HMMT25(Feb)~\cite{balunovic2026matharena}, HMMT25(Nov)~\cite{balunovic2026matharena}, HMMT26(Feb), and SMT25~\cite{he2025deepmath}. The main model comparisons report four benchmarks. The calibration ablations in Appendix~\ref{app:lambda_ablation} provide per-coefficient results on all eight benchmarks; these are sensitivity results rather than an eight-benchmark comparison of all baselines. For Code, we evaluate on TACO, Codeforces, and APPS; for instruction following (IF), we evaluate on Nemotron. Both Code and IF report Mean@8 (the average score over eight independently sampled responses). The Math and Code averages in Table~\ref{tab:math_code_performance} are arithmetic means over the benchmarks reported for each domain.



\begin{table}[h]
\centering
\renewcommand{\arraystretch}{1.1}
\caption{Training hyper-parameters of GRPO in math, code and instruction following RL.}
\begin{tabular}{lc}
\toprule
\textbf{Hyper-param.} & \textbf{Value} \\
\midrule
Train Batch Size     & 128 \\
Rollout $n$          & 8 \\
PPO Mini Batch Size  & 128 \\
Max Response Length  & 16{,}384 \\
Temperature          & 1.0 \\
Top-$p$              & 1.0 \\
LR                   & $1 \times 10^{-6}$ \\
KL Coefficient       & 0.0 \\
\bottomrule
\end{tabular}
\label{tab:grpo_hyperparams}
\end{table}

\begin{table}[h]
\centering
\caption{Training hyper-parameters of \methodAbb{} and OPD on math, code and instruction following domain.}
\begin{tabular}{l c}
\toprule
\textbf{Hyper-parameter} & \textbf{Value} \\
\midrule
Train Batch Size      & 256 \\
Rollout $n$           & 8 \\
PPO Mini Batch Size   & 1024 \\
Max Response Length   & 16{,}384 \\
Temperature           & 1.0 \\
Top-$p$               & 1.0 \\
LR                    & $1\times10^{-6}$ \\
\bottomrule
\end{tabular}

\label{tab:pust_hyperparams_math}
\end{table}




\begin{figure}[t]
\centering
\begin{subfigure}{0.48\linewidth}
    \centering
    \includegraphics[width=\linewidth]{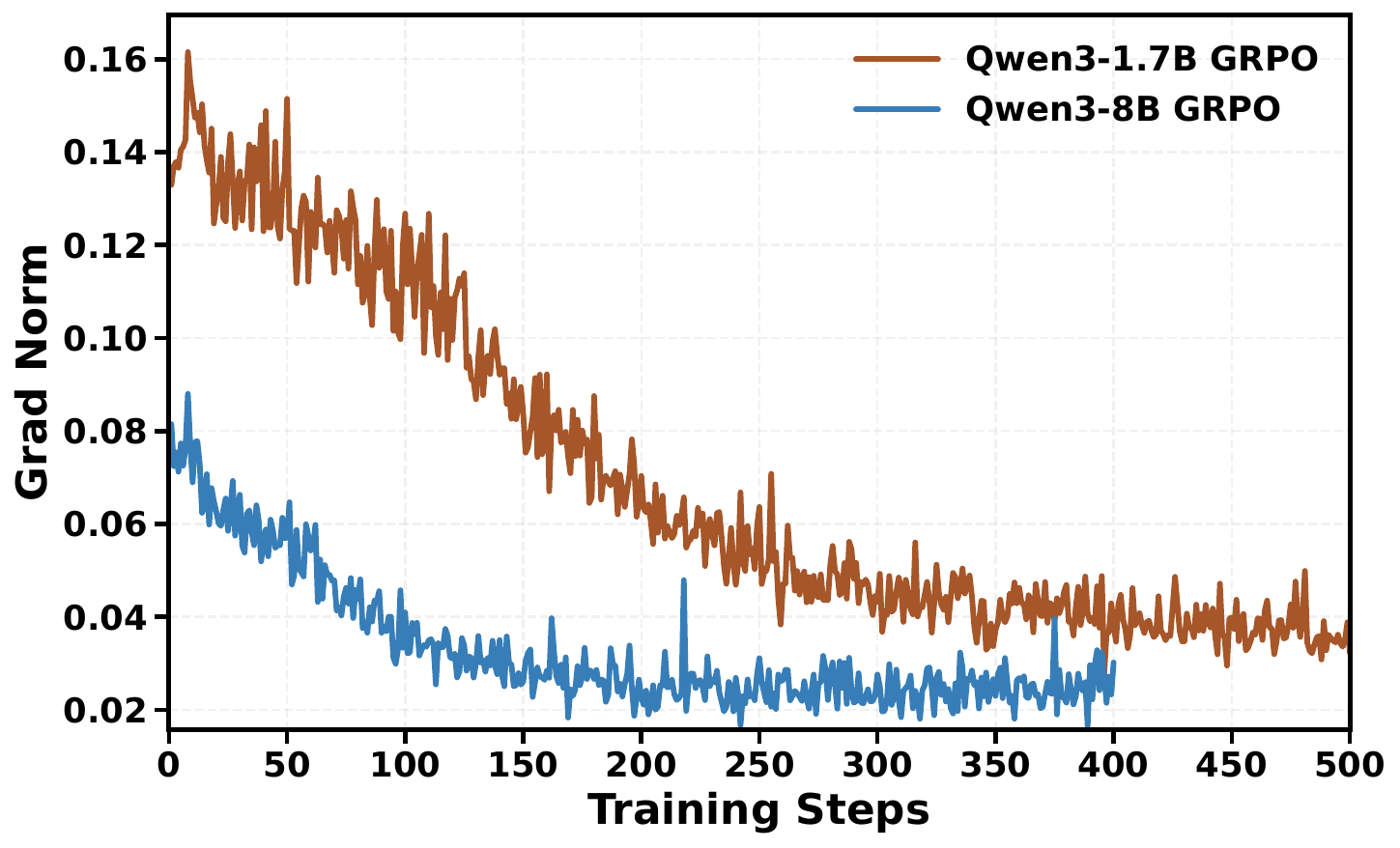}
    \caption{Training curve of gradient norm.}
    \label{fig:grad_norm}
\end{subfigure}%
\hfill
\begin{subfigure}{0.48\linewidth}
    \centering
    \includegraphics[width=\linewidth]{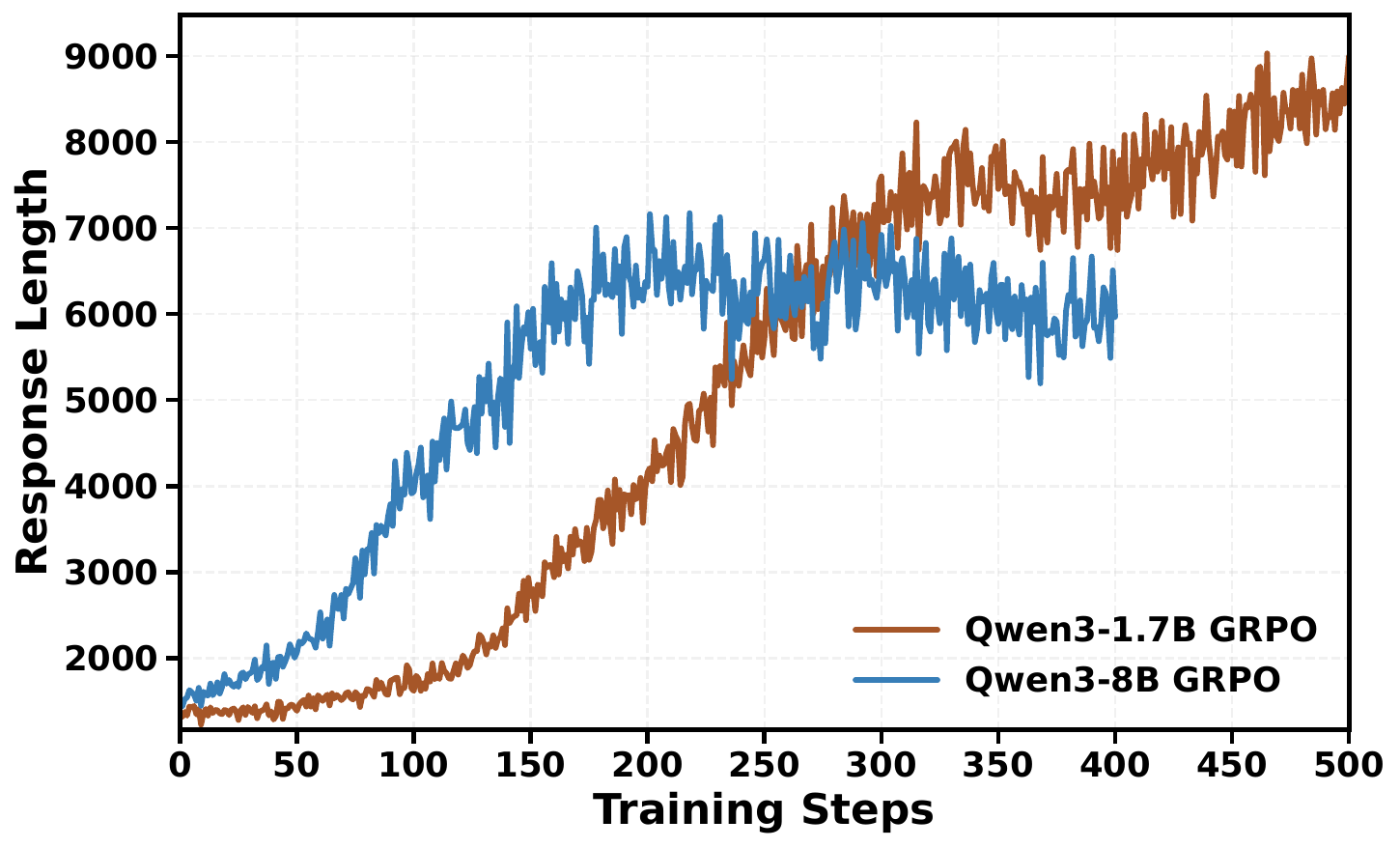}
    \caption{Training curve of response length.}
    \label{fig:grpo_response_length}
\end{subfigure}
\caption{
Left: \textbf{Gradient norm during GRPO training of the proxy and target reference expert.}
Right: \textbf{Response length during GRPO training of the proxy and target reference expert.}
}
\label{fig:grpo-training-dynamics}
\end{figure}

Proxy exploration uses the domain-specific training budgets described in Section~\ref{exp:efficiency}; additional Math runs vary the exploration budget to study signal quality in Section~\ref{sec:effectiveness}. Table~\ref{tab:grpo_hyperparams} summarizes the shared Math and Code GRPO hyper-parameters. The following curves provide diagnostic evidence about the evolution of the explored policies, rather than a separate stopping rule.

Figure~\ref{fig:grpo-training-dynamics} shows GRPO training dynamics for the Qwen3-1.7B proxy and the Qwen3-8B target reference expert. Gradient norms decline and fluctuate at lower levels later in training, while response lengths increase with model-dependent trajectories. These measurements describe optimization and generation behavior.

\subsection{Effect of the Calibration Coefficient $\lambda$}
\label{app:lambda_ablation}













\begin{table}[t]
\centering
\renewcommand{\arraystretch}{1.15}
\setlength{\tabcolsep}{3.5pt}

\caption{Mean@16 results on different math benchmarks of
\methodAbb{} (4B$\rightarrow$8B). Performance comparison across
calibration coefficients $\lambda$. The best results are highlighted
in \textbf{bold}.}
\label{tab:coefficient_tuning_4b_8b}

\resizebox{0.54\linewidth}{!}{%
\begin{tabular}{l ccccccc}
\toprule
\multicolumn{1}{c}{\multirow{2}{*}[-0.4ex]{\textbf{Benchmark}}}
& \multicolumn{7}{c}{\textbf{$\lambda$}} \\
\cmidrule(lr){2-8}
& \textbf{0.0}
& \textbf{0.5}
& \textbf{0.75}
& \textbf{0.95}
& \textbf{1.0}
& \textbf{1.1}
& \textbf{1.5} \\
\midrule
\textbf{AIME24}
& 0.0 & 57.3 & 61.3 & 60.8 & 62.5 & \textbf{63.5} & 60.4 \\

\textbf{AIME25}
& 0.0 & 41.9 & 50.6 & 50.2 & 52.7 & \textbf{52.9} & 48.3 \\

\textbf{CMIMC25}
& 0.0 & 8.8 & 27.8 & 29.2 & 28.4 & 29.5 & \textbf{31.9} \\

\textbf{AIME26}
& 0.0 & 45.6 & \textbf{56.9} & 56.7 & 56.7 & 55.6 & 52.3 \\

\textbf{HMMT25(Feb)}
& 0.0 & 23.8 & \textbf{34.2} & 31.9 & \textbf{34.2} & 32.1 & 27.9 \\

\textbf{HMMT25(Nov)}
& 0.0 & 26.7 & 37.7 & \textbf{41.0} & 40.4 & 40.8 & 37.5 \\

\textbf{SMT25}
& 0.0 & 12.3 & \textbf{52.5} & \textbf{52.5} & 50.9 & 52.0 & 19.8 \\

\textbf{HMMT26(Feb)}
& 0.0 & 30.4 & 32.2 & 33.3 & 33.5 & \textbf{34.9} & 31.8 \\

\textbf{AVG.}
& 0.0 & 30.9 & 44.1 & 44.5 & 44.9 & \textbf{45.2} & 38.7 \\
\bottomrule
\end{tabular}%
}
\end{table}

As described in Section~\ref{subsec:signal_transfer}, the calibration coefficient $\lambda$ balances proxy-signal absorption against deviation from the target's frozen base: a smaller $\lambda$ follows the proxy-induced direction more aggressively, while a larger $\lambda$ yields more conservative transfer. We sweep $\lambda$ for transfer from 4B and 1.7B proxies to an 8B target. The tables report Mean@16 on the eight Math benchmarks listed in Appendix~\ref{app:impl_details}.

\begin{figure}[t]
    \centering

    \begin{subfigure}{0.44\linewidth}
        \centering
        \includegraphics[width=\linewidth]
        {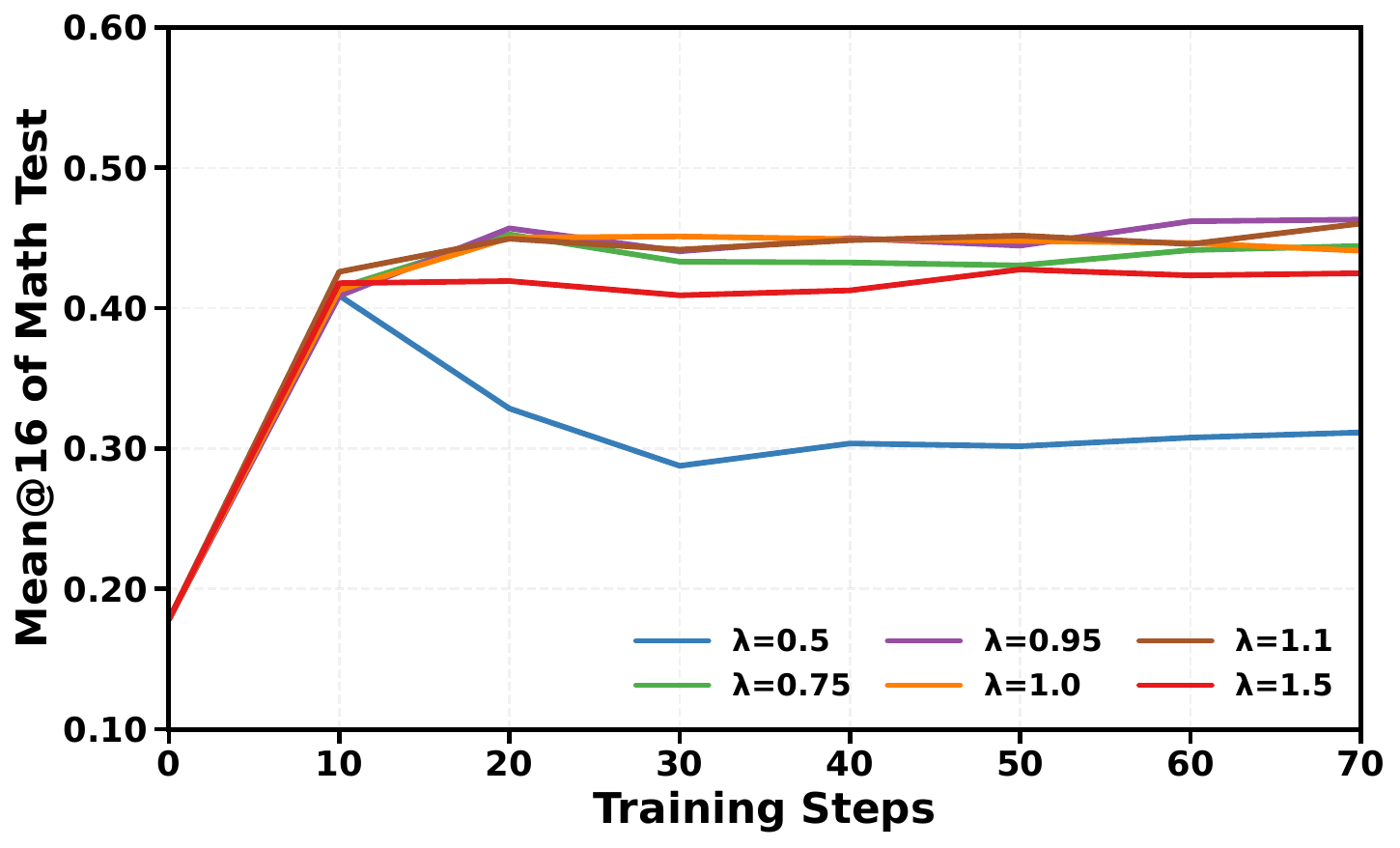}
        \caption{Math evaluation accuracy (Mean@16) under different calibration coefficients
        $\lambda$ (4B $\rightarrow$ 8B).}
        \label{fig:pupt_4b8b_val_acc}
    \end{subfigure}
    \hfill
    \begin{subfigure}{0.44\linewidth}
        \centering
        \includegraphics[width=\linewidth]
        {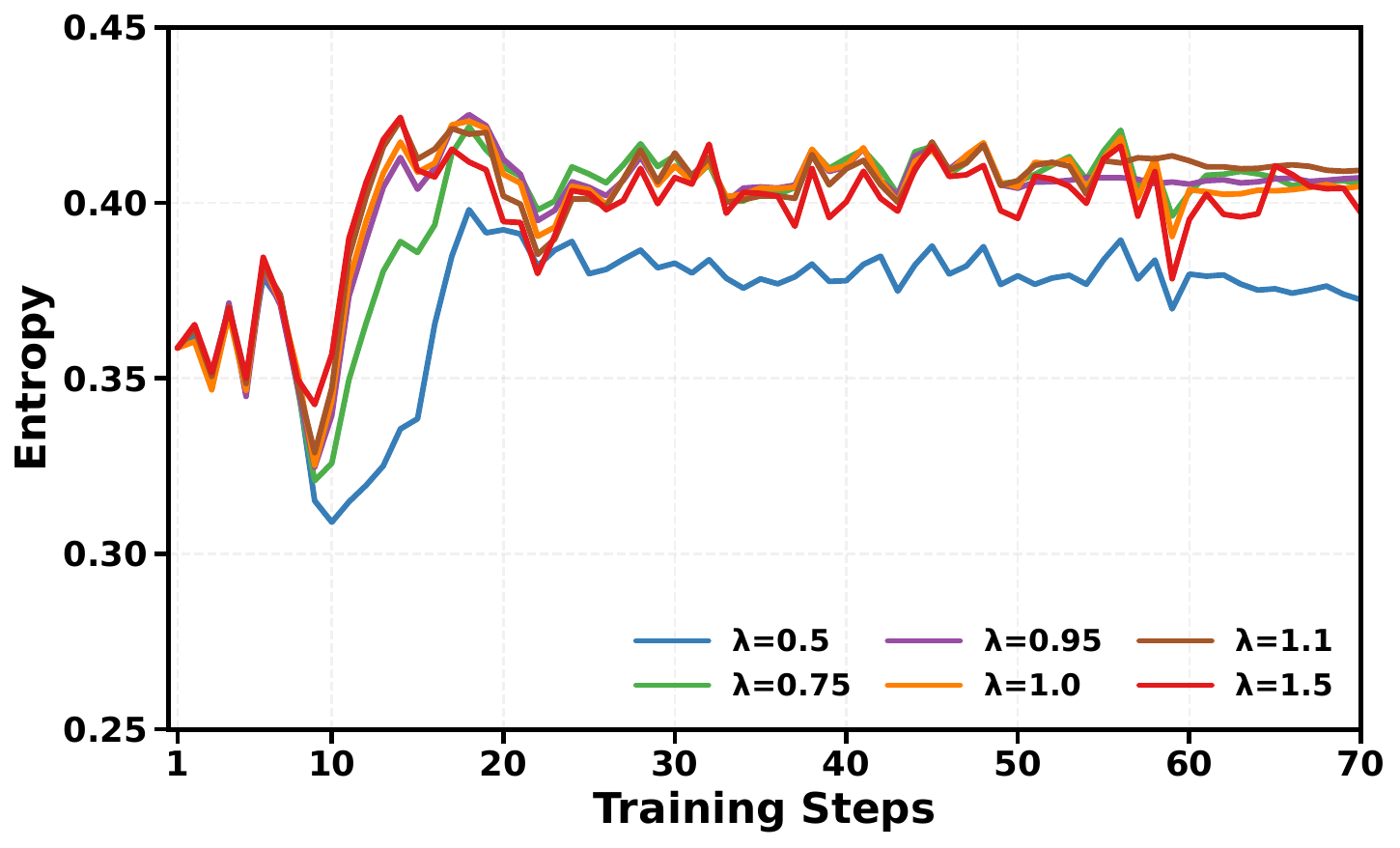}
        \caption{Conditioned entropy of target model under different
        $\lambda$ (4B $\rightarrow$ 8B).}
        \label{fig:pupt_4b8b_entropy}
    \end{subfigure}

    \medskip

    \begin{subfigure}{0.44\linewidth}
        \centering
        \includegraphics[width=\linewidth]
        {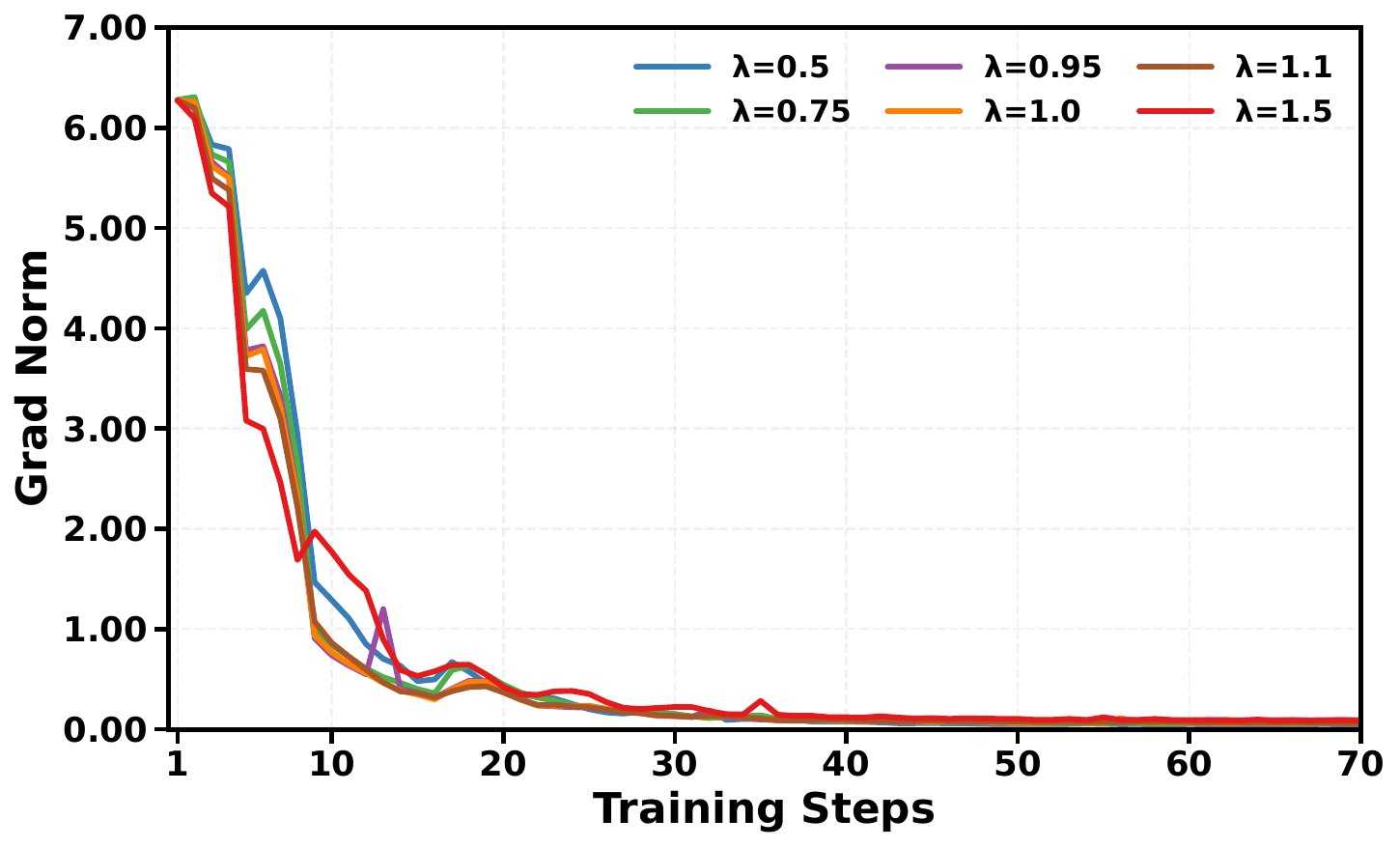}
        \caption{Gradient norm under different calibration coefficient
        $\lambda$ (4B $\rightarrow$ 8B).}
        \label{fig:pupt_4b8b_grad_norm}
    \end{subfigure}
    \hfill
    \begin{subfigure}{0.44\linewidth}
        \centering
        \includegraphics[width=\linewidth]
        {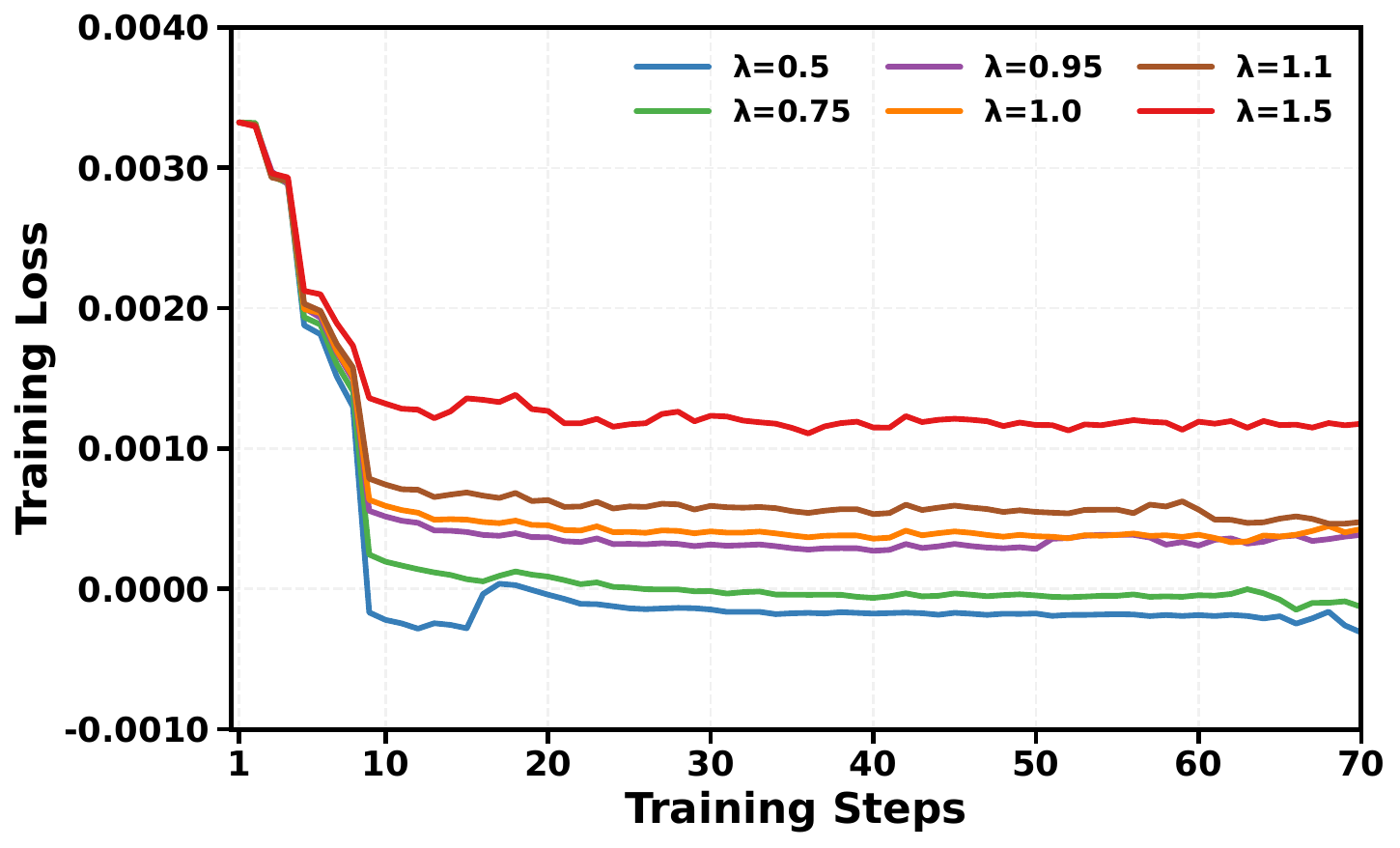}
        \caption{Training loss under different calibration coefficient
        $\lambda$ (4B $\rightarrow$ 8B).}
        \label{fig:pupt_4b8b_opd_loss}
    \end{subfigure}

    \caption{Effects of different calibration coefficients $\lambda$
    for the 4B $\rightarrow$ 8B setting.}
    \label{fig:pupt_4b8b_lambda}
    \vspace{-1em}
\end{figure}

\paragraph{4B$\rightarrow$8B transfer.} Panels (a) of Figure~\ref{fig:pupt_4b8b_lambda} show the training dynamics when the 4B proxy guides the 8B model, and Table~\ref{tab:coefficient_tuning_4b_8b} reports the corresponding benchmark scores. The positive-coefficient runs improve rapidly in the early steps, but the smallest positive coefficient ($\lambda=0.5$) is too aggressive: its evaluation accuracy peaks early and then degrades, and its entropy stays low, consistent with over-updating. The additional $\lambda=0$ ablation removes base-policy regularization and leads to complete performance collapse; the positive-$\lambda$ closed form in Eq.~\ref{eq:closed_form_target} does not apply to this ablation. Moderate coefficients ($\lambda\in\{0.75,0.95,1.0,1.1\}$) achieve higher average benchmark scores, while $\lambda=1.5$ is more conservative and remains effective despite a lower average. This suggests that a sufficient amount of calibration is needed to avoid over-following the fixed proxy direction.

\begin{table}[t]
\centering
\renewcommand{\arraystretch}{1.15}
\setlength{\tabcolsep}{4pt}

\caption{Mean@16 results on math benchmarks of
\methodAbb{} (1.7B$\rightarrow$8B). Performance comparison across
calibration coefficients $\lambda$. AVG. is the arithmetic mean of the
benchmark scores. All scores are rounded to one decimal place.
The best results are highlighted in \textbf{bold}.}
\label{tab:coefficient_tuning_1_7b_8b}

\resizebox{0.50\linewidth}{!}{%
\begin{tabular}{l cccccc}
\toprule
\multicolumn{1}{c}{\multirow{2}{*}[-0.4ex]{\textbf{Benchmark}}}
& \multicolumn{6}{c}{\textbf{$\lambda$}} \\
\cmidrule(lr){2-7}
& \textbf{0.5}
& \textbf{1.0}
& \textbf{1.25}
& \textbf{1.5}
& \textbf{1.75}
& \textbf{2.0} \\
\midrule
\textbf{AIME24}
& 0.0 & 41.0 & 50.8 & 50.8 & \textbf{51.7} & 47.7 \\

\textbf{AIME25}
& 0.0 & 31.9 & 37.9 & \textbf{42.1} & 39.2 & 36.9 \\

\textbf{CMIMC25}
& 0.0 & 16.4 & 22.5 & \textbf{26.4} & 25.5 & 25.0 \\

\textbf{AIME26}
& 0.0 & 32.3 & 43.1 & \textbf{46.3} & 43.1 & 36.9 \\

\textbf{HMMT25(Feb)}
& 0.0 & 13.5 & 19.6 & \textbf{22.9} & 22.5 & 21.9 \\

\textbf{HMMT25(Nov)}
& 0.0 & 25.0 & 29.2 & \textbf{30.6} & 30.0 & 29.4 \\

\textbf{SMT25}
& 0.0 & 35.9 & 42.9 & \textbf{48.2} & 44.1 & 42.2 \\

\textbf{HMMT26(Feb)}
& 0.0 & 24.1 & 27.7 & 28.6 & \textbf{29.6} & 29.0 \\

\textbf{AVG.}
& 0.0 & 27.5 & 34.2 & \textbf{37.0} & 35.7 & 33.6 \\
\bottomrule
\end{tabular}%
}
\end{table}

\begin{figure}
    \centering

    \begin{subfigure}{0.44\linewidth}
        \centering
        \includegraphics[width=\linewidth]
        {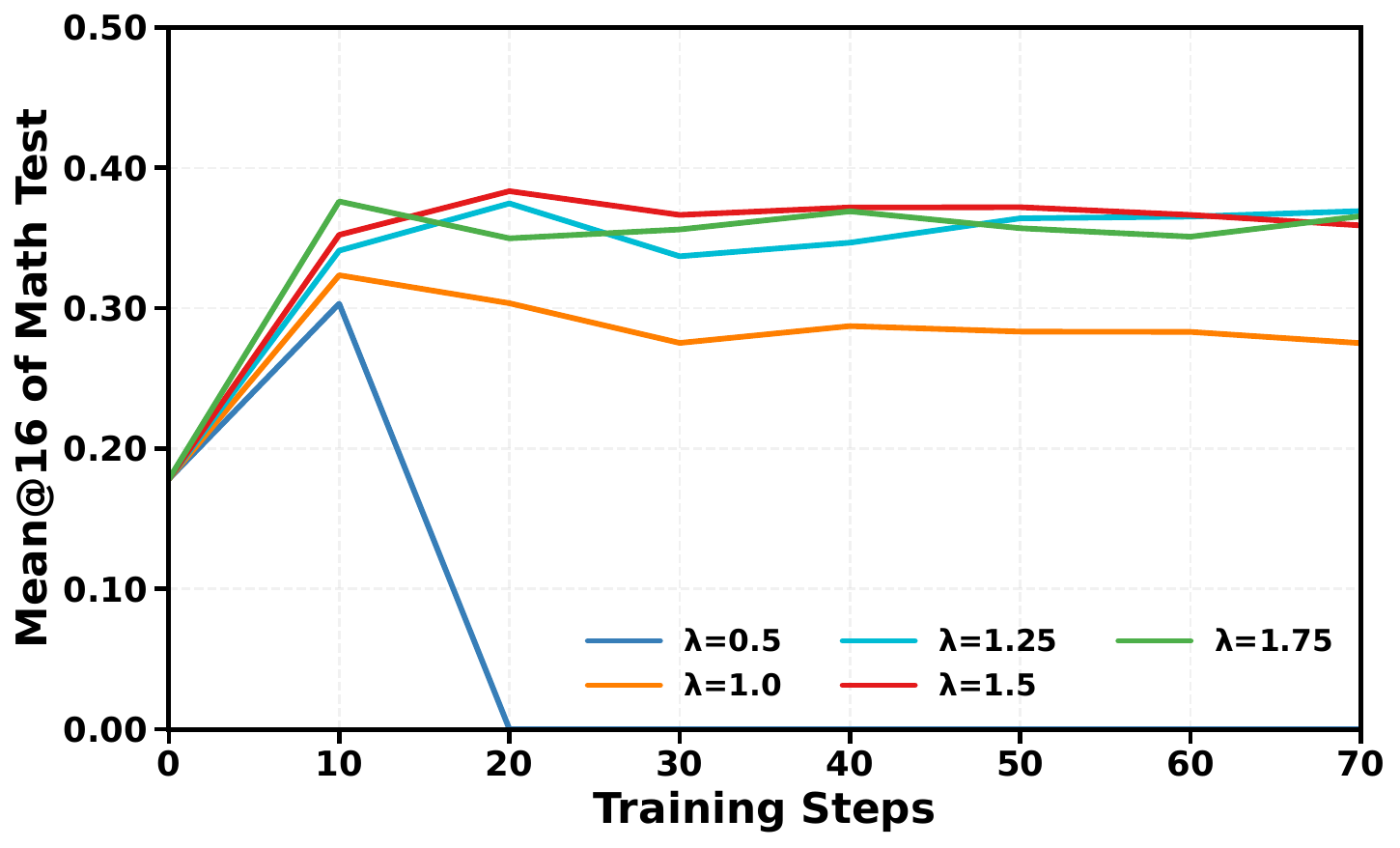}
        \caption{Math evaluation accuracy (Mean@16) under different calibration coefficients
        $\lambda$ (1.7B $\rightarrow$ 8B).}
        \label{fig:pupt_17b8b_val_acc}
    \end{subfigure}
    \hfill
    \begin{subfigure}{0.44\linewidth}
        \centering
        \includegraphics[width=\linewidth]
        {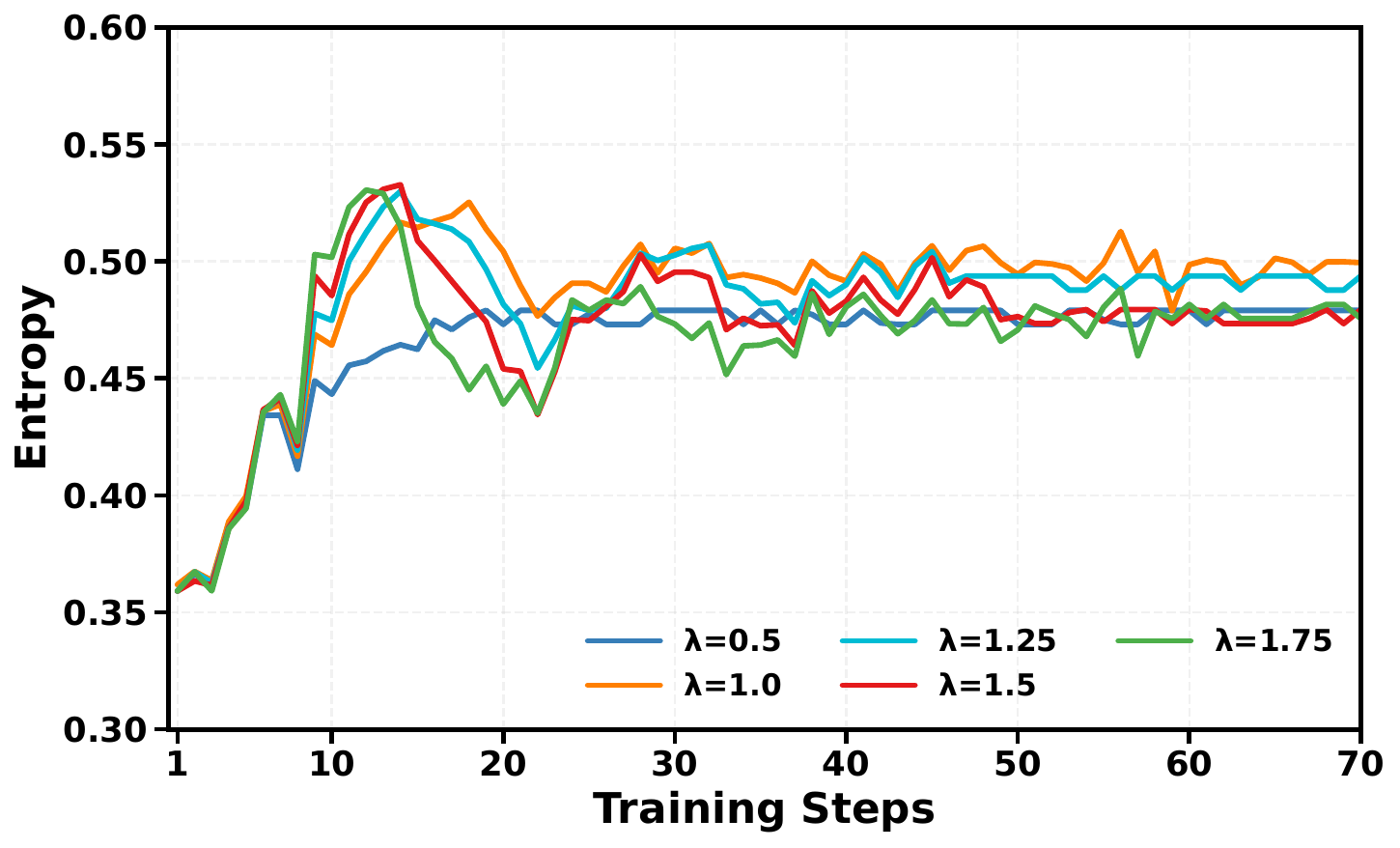}
        \caption{Conditioned entropy of target model under different
        $\lambda$ (1.7B $\rightarrow$ 8B).}
        \label{fig:pupt_17b8b_entropy}
    \end{subfigure}

    \medskip

    \begin{subfigure}{0.44\linewidth}
        \centering
        \includegraphics[width=\linewidth]
        {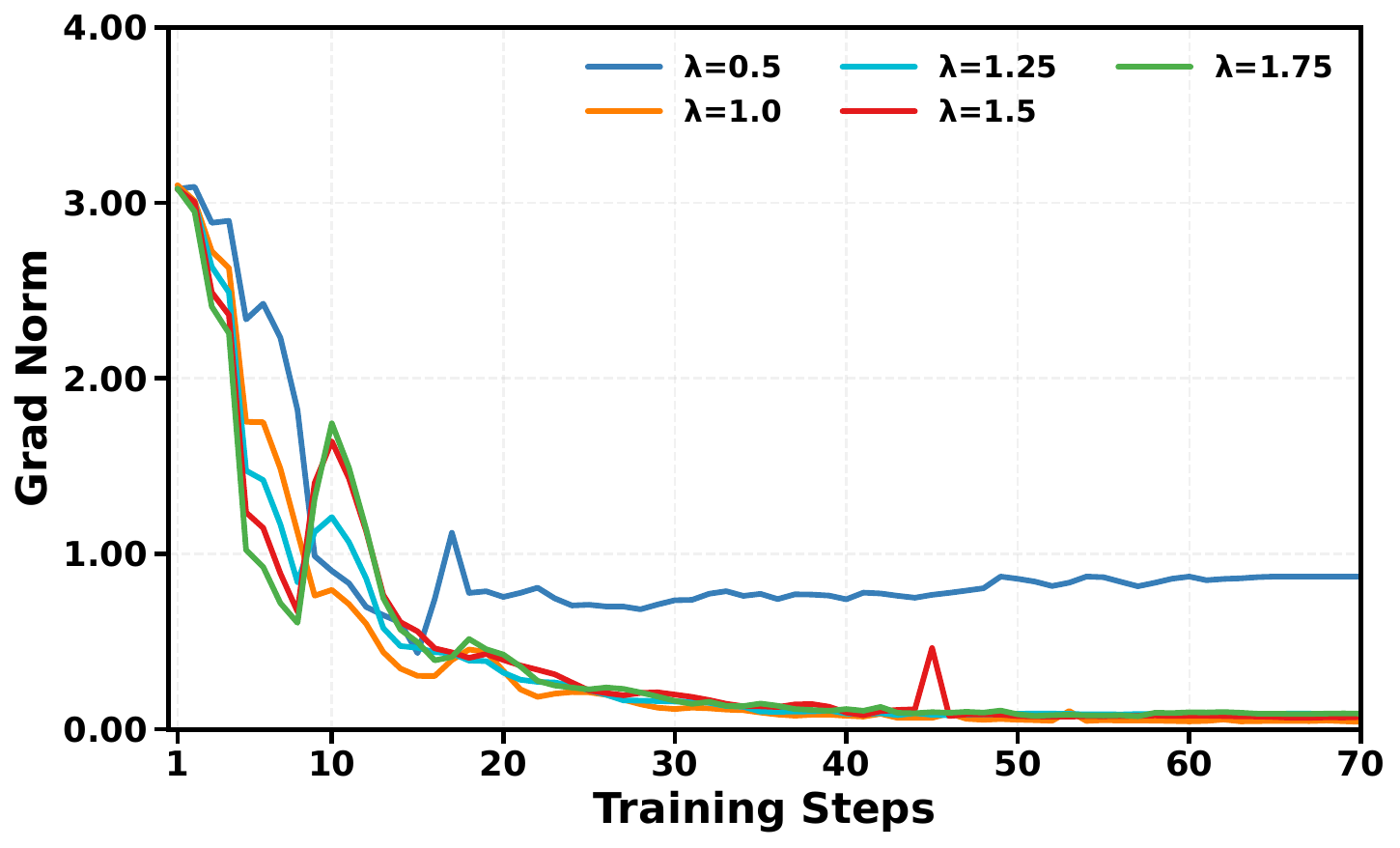}
        \caption{Gradient norm under different calibration coefficient
        $\lambda$ (1.7B $\rightarrow$ 8B).}
        \label{fig:pupt_17b8b_grad_norm}
    \end{subfigure}
    \hfill
    \begin{subfigure}{0.44\linewidth}
        \centering
        \includegraphics[width=\linewidth]
        {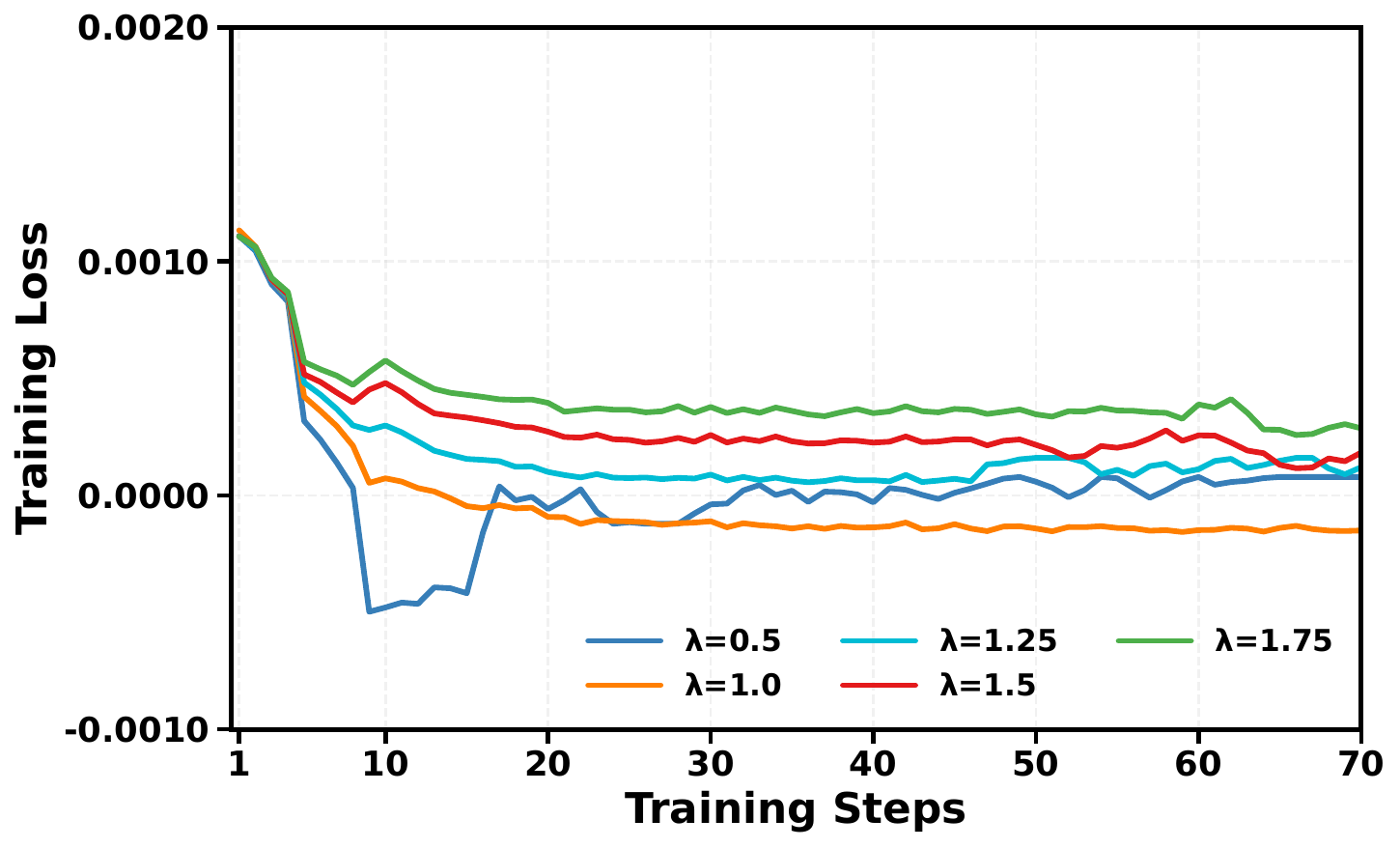}
        \caption{Training loss under different calibration coefficient
        $\lambda$ (1.7B $\rightarrow$ 8B).}
        \label{fig:pupt_17b8b_opd_loss}
    \end{subfigure}

    \caption{Effects of different calibration coefficients $\lambda$
    for the 1.7B $\rightarrow$ 8B setting.}
    \label{fig:pupt_17b8b_lambda}
\end{figure}

\paragraph{1.7B$\rightarrow$8B transfer.} Panels (b) of Figures~\ref{fig:pupt_17b8b_lambda} show the training dynamics for the smaller 1.7B proxy, with benchmark scores reported in Table~\ref{tab:coefficient_tuning_1_7b_8b}. Average benchmark accuracy is lower than in the 4B proxy setting. The intermediate coefficient $\lambda=1.5$ achieves the highest eight-benchmark average, whereas the more conservative settings $\lambda\in\{1.75,2.0\}$ yield lower averages. At $\lambda=0.5$, performance collapses. These results indicate that the smaller proxy benefits from stronger regularization than the 4B proxy in this setting, but increasing conservatism beyond the best observed coefficient does not improve performance.

\end{document}